\documentclass[letterpaper]{article} 
\usepackage{aaai2027}  
\usepackage[hyphens]{url}  
\usepackage{graphicx} 
\usepackage{natbib}  
\usepackage{caption} 
\usepackage{subcaption}

\usepackage{algorithm}
\usepackage{algorithmic}

\usepackage{newfloat}
\usepackage{xspace}
\usepackage{listings}
\DeclareCaptionStyle{ruled}{labelfont=normalfont,labelsep=colon,strut=off} 
\floatstyle{ruled}
\newfloat{listing}{tb}{lst}{}
\floatname{listing}{Listing}

\usepackage{booktabs}

\usepackage{amsmath}
\usepackage{amssymb}

\nocopyright 

\usepackage{colortbl}
\usepackage{multirow}
\usepackage[table,dvipsnames]{xcolor}
\usepackage{siunitx}

\newcommand{\mosaik}{\textsc{MOSAIK}\xspace}

\title{MOSAIK: Multi-Patch Content-Aware Spatial Allocation \\ of Image Tokens for Efficient Generation}
\author {
    Mohammadreza Hami\textsuperscript{*}, 
    ~Mohammadreza Samadi\textsuperscript{*}, 
    ~Chao Gao, 
    ~Negar Hassanpour
}
\affiliations {
    Huawei Technologies Canada\\
    mohammadreza.hami@h-partners.com\\
    \{mohammadreza.samadi1, chao.gao4, negar.hassanpour2\}@huawei.com\\
    \textsuperscript{*} Equal contribution
}

\begin{document}

\maketitle

\begin{abstract}
Pixel-space diffusion models avoid the reconstruction ceiling of latent diffusion models by generating directly in image space.
However, their substantially higher token count makes generation expensive due to the quadratic complexity of self-attention.
Several existing efficiency methods reduce this cost by using larger patches at selected denoising steps, thereby representing the image with fewer tokens.
Yet, each step still uses a single patch size uniformly across the entire image,
overlooking that different regions suffer different fidelity losses when coarsened.
We introduce MOSAIK, a damage-guided framework that varies patch size across regions and denoising steps.
MOSAIK adapts the PixelDiT backbone to generate arbitrary heterogeneous patch layouts, and
a lightweight predictor uses intermediate denoising features to estimate the fidelity loss caused by coarsening each region.
Given a token budget, our damage-guided layout predictor assigns fine patches to sensitive regions and coarse patches elsewhere.
Remarkably, while reducing FLOPs by 70\% and token count by 83\%, MOSAIK matches the full-compute PixelDiT on GenEval and its \hbox{DPG-Bench} score drops by only 1.0 point.
Compared to diverse efficiency paradigms, including temporal patch scheduling and feature caching, our approach delivers highly competitive performance at moderate budgets and consistently outperforms these baselines in highly constrained compute regimes.
\end{abstract}


\section{Introduction}
\label{sec:intro}

Diffusion models~\citep{ho2020ddpm,dit2023,lipman2023flow} have become a leading paradigm for high-fidelity image synthesis, generating complex images through an iterative denoising process often parameterized by large transformers~\citep{vaswani2017attention}. To contain the computational cost of this process, most diffusion models operate in a compressed latent space produced by a pretrained variational autoencoder~(VAE)~\citep{kingma2014auto,rezende2014stochastic}, as in latent diffusion models~(LDMs)~\citep{ldm2022,dcae2024}. This efficiency, however, comes with an inherent reconstruction ceiling: information discarded by the VAE, particularly fine-grained and high-frequency spatial details, cannot be recovered by the diffusion model.

\begin{figure}[!t]
    \centering
    \includegraphics[width=\linewidth]{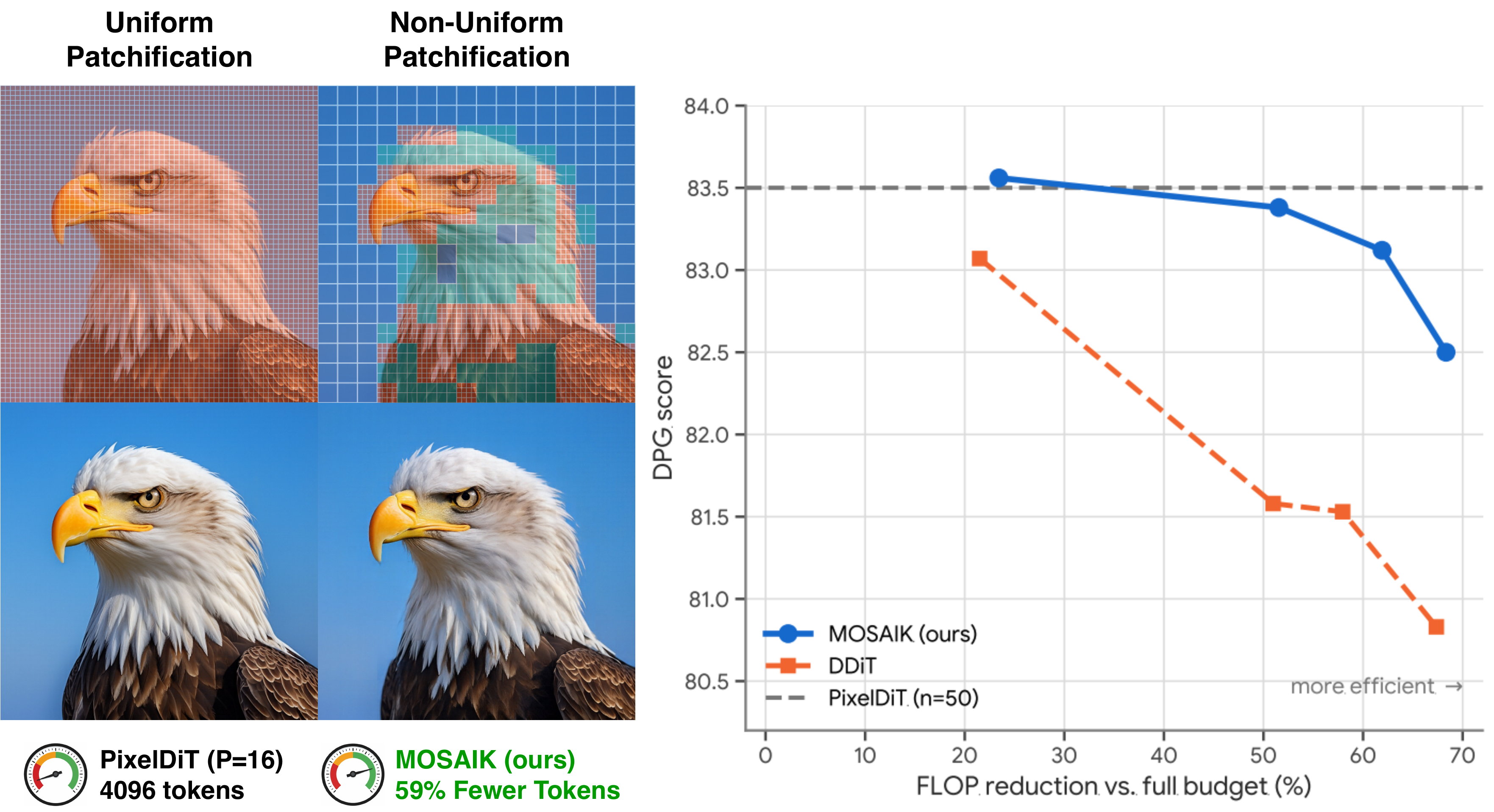}
    \caption{\textbf{Mixed-patch generation is highly efficient.} \emph{Left:} \mosaik{} matches uniform patch-16 quality with a $50\%$ reduction in compute (FLOPs) by allocating smaller patches to proper regions. \emph{Right:} Unlike temporal scheduling or step reduction, \mosaik{} sustains image quality as compute decreases.}
\label{fig:motivation}
\end{figure}
\begin{figure*}[!t]
    \centering
    \includegraphics[width=\linewidth]{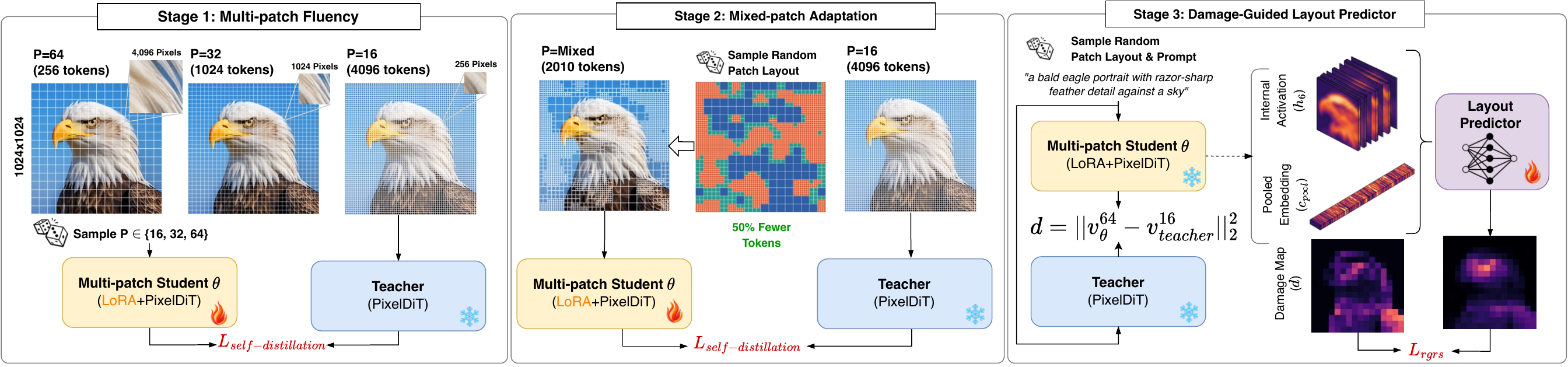}
    \caption{\textbf{\textsc{MOSAIK} training pipeline.} The model is trained in three stages via self-distillation from a frozen PixelDiT teacher. \textbf{Stage 1} trains a model to render all uniform patch sizes ($P \in \{16, 32, 64\}$). \textbf{Stage 2} adapts the model to process arbitrary mixed-patch layouts in a single forward pass, yielding a mixed-patch generator. \textbf{Stage 3} trains an offline layout predictor to regress a per-region coarsening damage map from intermediate activations ($\mathbf{h}_6$) and prompt embeddings ($\mathbf{c}_{\text{pool}}$).}
\label{fig:overview}
\end{figure*}

Recent pixel-space diffusion models~(PDMs)~\citep{jit2025,pixeldit2025,deco2025,dip2025,hyperdit2026} avoid this ceiling by generating directly in image space. However, this exposes the denoising transformer to substantially more image tokens. At a fixed image resolution, each patch is represented by one token. As shown in Fig.~\ref{fig:overview}~(left), larger patches reduce the token count by encoding a broader spatial area within each token. A $p16$ token represents 256 pixels, whereas a $p64$ token represents 4096 pixels using the same embedding dimension. This stronger spatial aggregation improves efficiency but can discard local structure and sharp boundaries, while smaller patches preserve these details and support higher visual fidelity.

Existing efficiency methods navigate this trade-off by varying the patch size across denoising timesteps~\citep{ddit2026,flexidit2025,ppflow2026} or network depth~\citep{mpdit2026}. Nevertheless, at any given timestep and layer, they apply a spatially uniform patch layout across the entire image. 
This subjects every image region to the same degree of spatial compression, 
creating either an \textit{efficiency} concern by overspending computation on simple regions or a \textit{quality} concern by 
constraining representational capacity in sensitive regions. This observation motivates \emph{spatially adaptive patchification}: retaining small patches where coarsening is harmful while assigning larger patches elsewhere to reduce computation without sacrificing visual quality, as illustrated in Fig.~\ref{fig:motivation}. 

\begin{figure*}[t!]
    \centering
    \includegraphics[width=\linewidth]{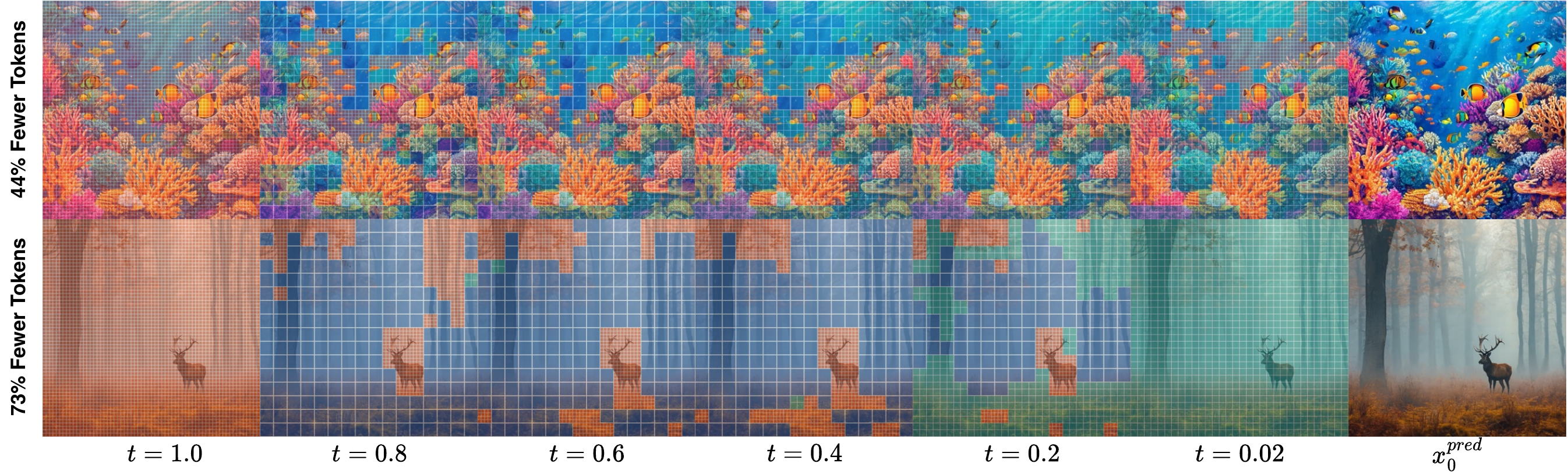}
    \caption{\textbf{Damage-guided patch layout across denoising.} Per-region patch size ($16$/$32$/$64$\,px; \textcolor{orange}{orange}/\textcolor{teal}{teal}/\textcolor{blue}{blue}) set from the predicted coarsening damage at each step, over the running estimate ($x_0^{\text{pred}}$). Fine patches concentrate on high-damage regions, coarse patches on low-damage ones.}
\label{fig:patch_layout_progress}
\end{figure*}

\paragraph{Contributions.}
We introduce \mosaik{}~(\textbf{M}ulti-patch c\textbf{O}ntent-aware \textbf{S}patial \textbf{A}llocation of \textbf{I}mage to\textbf{K}ens), a framework that employs spatially heterogeneous patch layouts to reduce computation while preserving fine-grained visual details. \mosaik{} first equips a pretrained diffusion model to operate at multiple patch sizes~(\S\ref{sec:stage1}) and then trains it to reliably execute arbitrary heterogeneous layouts~(\S\ref{sec:stage2}). We further formulate patch allocation as a damage-prediction problem and train a lightweight predictor to estimate the local degradation caused by coarsening each region
~(\S\ref{sec:stage3}; Fig.~\ref{fig:patch_layout_progress}). At inference, layout predictor dynamically allocates tokens up to a user-specified budget at each denoising step, assigning finer patches to regions where coarsening is predicted to cause the greatest degradation, without retraining. 

Through controlled ablations and extensive experiments, we show that damage-guided patchification consistently improves the quality/computation trade-off over competing methods across multiple benchmarks and efficiency budgets. Specifically, \mosaik{} consistently maintains stable performance across aggressive FLOPs reductions, matching full-resolution quality at a $60\%$ reduction and sustaining performance at an $80\%$ reduction when integrated with caching methods.

\section{Related Work}
\label{sec:related}

\paragraph{From latent to pixel space.}
Latent diffusion models~(LDMs)~\cite{ldm2022,dit2023} have significantly reduced the computational burden of iterative denoising by operating within the compressed latent space of a pretrained VAE~\citep{kingma2014auto,rezende2014stochastic}. However, this efficiency imposes a hard fidelity ceiling because aggressive spatial compression inevitably removes fine-grained information~\cite{dcae2024}. Pixel-space diffusion models~(PDMs)~\cite{jit2025,deco2025,dip2025,pixeldit2025,hyperdit2026} bypass this limitation by generating directly in pixel space, replacing the representational-capacity bottleneck of VAE with a computational-scalability bottleneck of PDMs: 
modeling raw pixels substantially increases the token count processed by the transformer at every denoising step. 

Recent architectures reduce the cost of uniformly fine tokenization while retaining fixed spatial grids. DiP~\cite{dip2025} uses a coarse backbone for efficient global processing and adds a dedicated branch to recover fine details. DeCo~\cite{deco2025} separates the image into frequency bands and processes them through distinct pathways. PixelDiT~\cite{pixeldit2025} combines patch-level representations for global structure with pixel-level representations for local detail. HyperDiT~\cite{hyperdit2026} retains selected fine tokens and connects them to semantic anchors in the coarse representation. Crucially, across all these methods, the patch size is fixed in advance and applied uniformly across the image, without accounting for spatial differences in sensitivity to coarsening.

\paragraph{Adaptive computation in diffusion transformers.}
Recent literature explores adaptive computation along three primary axes: time, depth, and token count. DDiT~\cite{ddit2026} uses a test-time scheduler based on finite differences of the latent trajectory to select a global patch size at each denoising step. In contrast, PPFlow~\cite{ppflow2026} partitions the denoising trajectory into predefined stages, using larger patches at high noise levels and smaller patches in later stages. Despite this difference between adaptive and fixed temporal scheduling, both methods apply a single patch size across all spatial regions at each step. Unlike these approaches, which treat spatial uniformity as a fixed constraint, our framework recognizes that visual information density is inherently heterogeneous, establishing the need for dynamic, per-region spatial allocation at every denoising step. 

Along the depth axis, MPDiT~\cite{mpdit2026} applies larger patches in early transformer blocks to capture global context and finer patches in later blocks for localized refinement, yet this schedule remains input-agnostic. Other token-reduction strategies merge, prune, or adapt token widths inside the network~\cite{tome2023,atedm2024,dydit2025,diffcr2025}. Ultimately, these depth-level and token-level methods optimize the computation of a static initial token set, rather than dynamically scaling the underlying pixel-to-token representation itself.

\paragraph{Content-aware spatial allocation.}
Motivated by the spatial heterogeneity of natural images, a line of research adapts patch sizes to image content. In discriminative tasks, mixed-scale tokenization and adaptive patching process homogeneous regions with larger patches while retaining smaller patches where fine spatial information is important for recognition~\cite{msvit2023,apt2026}. In generative modeling, however, applying dynamic spatial allocation within LDM faces a structural bottleneck: the standard VAE performs a uniform, offline compression that prematurely consumes spatial redundancy and obscures the local signals needed to efficiently determine where coarsening is safe. D$^2$iT~\cite{d2it2025} applies region-dependent granularity but necessitates a fully retrained VAE. Foveated Diffusion~\cite{foveated2026} determines the spatial patch granularity from an externally provided gaze mask rather than directly inferring it from image content. Similarly, PatchScaler~\cite{patchscaler2025} assigns region groups different diffusion starting points and sampling-step counts based on reconstruction difficulty, rather than varying their patch sizes.

Concurrent to our work, DynaPatch~\cite{dynapatch2026} performs dynamic patchification for video diffusion. Operating in the highly compressed latent space of a 3D-VAE~\cite{adaptok2026}, it finds that heuristic content statistics are insufficient for predicting effective patch layouts in latent space and therefore jointly optimizes its router with the generative objective. Our findings reveal that PDMs naturally resolve these bottlenecks. Because there is no uniform VAE pre-compression, a higher capacity of spatial redundancy remains available for the diffusion model to dynamically utilize~(see Appendix). Furthermore, intermediate diffusion features in PDMs remain directly aligned with spatial image regions, providing a transparent signal of where coarsening is safe. We use these features to train the layout predictor separately from the generator, allowing it to estimate where coarse processing would cause the greatest degradation. This enables dynamic spatial allocation without the end-to-end joint optimization required by latent-space routing methods.

\section{Preliminaries}
\label{sec:prelim}

\subsection{PixelDiT as the Diffusion Backbone}
We build on PixelDiT~\citep{pixeldit2025}, a pixel-space diffusion model for $1024 \times 1024$ image generation. Given a noisy image $x_t$, noise level $\sigma$, and text condition $y$, the model predicts the velocity field $v_\theta(x_t,\sigma,y)$. PixelDiT partitions an $H \times W$ image into non-overlapping $P \times P$ patches, producing $HW/P^2$ image tokens. The pretrained model exclusively uses a uniform patch size of $16$.
Throughout this work, the \emph{teacher} denotes the original PixelDiT, whose prediction $v_{\mathrm{teacher}}^{16}$ serves as the fidelity reference.

PixelDiT consists of a patch-level DiT and a lightweight pixel-level decoder~(PiT). The DiT can process different patch sizes, while the PiT operates on a fixed uniform $p16$ grid to preserve high-frequency details. Patch-size dependence is confined to the tokenizer, output projection head, and positional embeddings; the internal DiT blocks operate on the resulting token sequence. This separation enables us to introduce dynamic patchification without modifying the PiT.

\subsection{Nomenclature}
We use the following terminology throughout the paper:

\noindent \textbf{Region.}
We partition each $1024 \times 1024$ image into a $16 \times 16$ grid of non-overlapping $64 \times 64$ pixel regions.
We denote the set of all $256$ regions by $\mathcal{R}$.
Each region is the smallest unit to which we independently assign a patch size.

\noindent \textbf{Patch size.}
We support three patch sizes,
$\mathcal{P}=\{16,32,64\}$,
denoted by $p16$,
$p32$,
and $p64$.
Within a $64 \times 64$ region,
these patch sizes produce $16$,
$4$,
and $1$ token,
respectively.

\noindent \textbf{Layout.}
A layout specifies the patch size assigned to every region.
Formally,
a layout is a mapping
$
\ell:\mathcal{R}\rightarrow\mathcal{P},
$
where $\ell(r)$ denotes the patch size assigned to region $r$.
A layout is \emph{uniform} if all regions use the same patch size,
and \emph{heterogeneous} otherwise.
We denote the uniform $p16$,
$p32$,
and $p64$ layouts by $\ell^{16}$,
$\ell^{32}$,
and $\ell^{64}$,
respectively.

\noindent \textbf{Coarsening and refinement.}
Changing a region from a smaller to a larger patch size is called \emph{coarsening},
and the vice-versa is called \emph{refinement}.
\section{Method}
\label{sec:method}

The core principle of \mosaik{} is to decouple the execution of heterogeneous patch layouts from the policy that determines their spatial allocation. We achieve this through a three-stage~training pipeline: aligning the multi-patch generator with the teacher's high-resolution prior across different uniform patch sizes~(\S\ref{sec:stage1}), extending 
the generation capability across spatially heterogeneous patch layouts
~(\S\ref{sec:stage2}), and training a standalone predictor that estimates the local degradation incurred by coarsening and dictates token routing at inference~(\S\ref{sec:stage3}).
Figure ~\ref{fig:overview} illustrates these three stages.

\subsection{Stage~1: Multi-Patch Fluency}
\label{sec:stage1}

\paragraph{Architectural Modifications.} 
To induce multi-patch fluency without catastrophically forgetting the original generative prior, we freeze the backbone weights and inject Low-Rank Adaptation (LoRA)~\citep{hu2022lora} modules into the DiT blocks. We also introduce separate tokenizers for $p32$ and $p64$, allowing each patch size to learn an appropriate input projection without modifying the pretrained $p16$ tokenizer. Moreover, we add a learned patch-size embedding to each token, similar to time embeddings, allowing the model to identify the patch size from which the token originates. To bridge the spatial mismatch with the fixed uniform $p16$ PiT, we introduce a learned pixel-shuffle conditioning upsampler that lifts the coarse DiT representations back onto the uniform $p16$ grid the PiT requires.

\paragraph{Training Strategy.} To prevent the spatial compression from degrading generation quality, we force the manifold of these newly introduced modules to closely match the high-fidelity prior of the teacher. At each training step, given a sampled noisy image $x_t$, noise level $\sigma$, text prompt $y$, and a uniform patch size $P \in \{16,32,64\}$, the multi-patch generator parameterized by $\theta$ is optimized via self-distillation to match the velocity predictions of the frozen teacher:

\begin{equation}
\mathcal{L}_{\text{self-distill}}
=
\mathbb{E}_{x_t,\sigma,y,P}
\left[
\left\|
v_\theta^{P}(x_t,\sigma,y)
-
v_{\mathrm{teacher}}^{16}(x_t,\sigma,y)
\right\|_2^2
\right].
\label{eq:distill}
\end{equation}

\subsection{Stage~2: Mixed-Patch Adaptation}
\label{sec:stage2}

\paragraph{Positional Embedding.}
Under heterogeneous patchification, $p16$, $p32$, and $p64$ tokens coexist within a single image. Because these tokens cover different spatial extents, they no longer lie on the uniform token grid assumed by the pretrained Rotary Positional Embeddings~(RoPE)~\citep{su2024roformer}. Therefore, assigning positions by sequence order would no longer reflect their actual locations in the image. We address this by anchoring each token to the geometric center of its patch in the normalized $p16$ coordinate frame. For example, a $p32$ token receives the center coordinate of the $2\times2$ block of $p16$ positions that it replaces. This layout-agnostic remapping seamlessly preserves the original positional geometry without requiring architectural changes.

\paragraph{Training Strategy.}
We train the generator on randomly sampled heterogeneous patch layouts, deliberately decoupling the spatial arrangement of patch sizes from the image content~(see Fig.~\ref{fig:overview}~(middle)). In each iteration, with probability 0.6, we apply a mixed layout spanning a broad range of token budgets, and with probability 0.4, we apply a uniform layout to preserve performance on the uniform configurations learned in Stage~1.
By retaining the objective in Equation~\eqref{eq:distill} and matching $v^{16}_{\text{teacher}}$ across all configurations, we force the generator to reliably execute arbitrary layouts regardless of how they align with the underlying visual features.

\subsection{Stage~3: Damage-Guided Layout Predictor}
\label{sec:stage3}

One of our key insights is that the degradation caused by spatial coarsening provides a direct allocation signal: regions that are more sensitive to larger patches should receive finer resolution. Rather than learning this behavior indirectly through the generative objective, we explicitly formulate dynamic patch allocation as a damage-prediction problem. At denoising step $t$, we compute the squared difference between the prediction of the multi-patch generator under a uniform $p64$ layout and that of the frozen $p16$ teacher. We then average-pool this error over the non-overlapping $64\times64$ regions to obtain the target damage map:
\begin{equation}
\tilde{d}^t
=
\mathrm{AvgPool}
\left(
\left\|
v^{64}_{\theta}(x_t,\sigma,y)
-
v^{16}_{\mathrm{teacher}}(x_t,\sigma,y)
\right\|_2^2
\right).
\label{eq:damage}
\end{equation}

The damage assigned to region $r$ is given by the corresponding entry of this map, denoted $\tilde{d}_r^t$. A larger value of $\tilde{d}_r^t$ indicates that uniform $p64$ processing causes greater deviation from the fine-grained teacher in region $r$. Such regions are assigned finer patches, while regions with lower damage remain coarse. Exact evaluation of Equation~\eqref{eq:damage} at inference requires a secondary forward pass, which negates the computational savings of adaptive patchification. Instead, we train a lightweight convolutional network $f_\phi$ with FiLM conditioning~\cite{perez2018film} to regress the damage map directly from intermediate backbone features. This design is motivated by representation-alignment works such as RePA~\citep{repa2024} that found intermediate denoising features encode meaningful visual information.

As shown in Fig.~\ref{fig:overview} (right), using an offline dataset of trajectories, the predictor learns to estimate the damage map $\tilde{d^t}$ from the DiT block-6 activations $\mathbf{h}_6^{t-1}$, conditioned on noise level $\sigma$ and pooled text embeddings $\mathbf{c}_{\text{pool}}$, via a Huber loss, which we employ to robustly handle extreme damage outliers while maintaining smooth convergence for minor prediction errors:
\begin{equation}
\mathcal{L}_{\text{rgrs}} = \mathbb{E}\left[\,\mathrm{Huber}\big(f_\phi(\mathbf{h}_6^{t-1}, \sigma, \mathbf{c}_{\text{pool}}),\; \tilde{d^t}\,\big)\right].
\label{eq:rgrs}
\end{equation}

\begin{figure*}[t]
\centering
\includegraphics[width=0.875\linewidth]{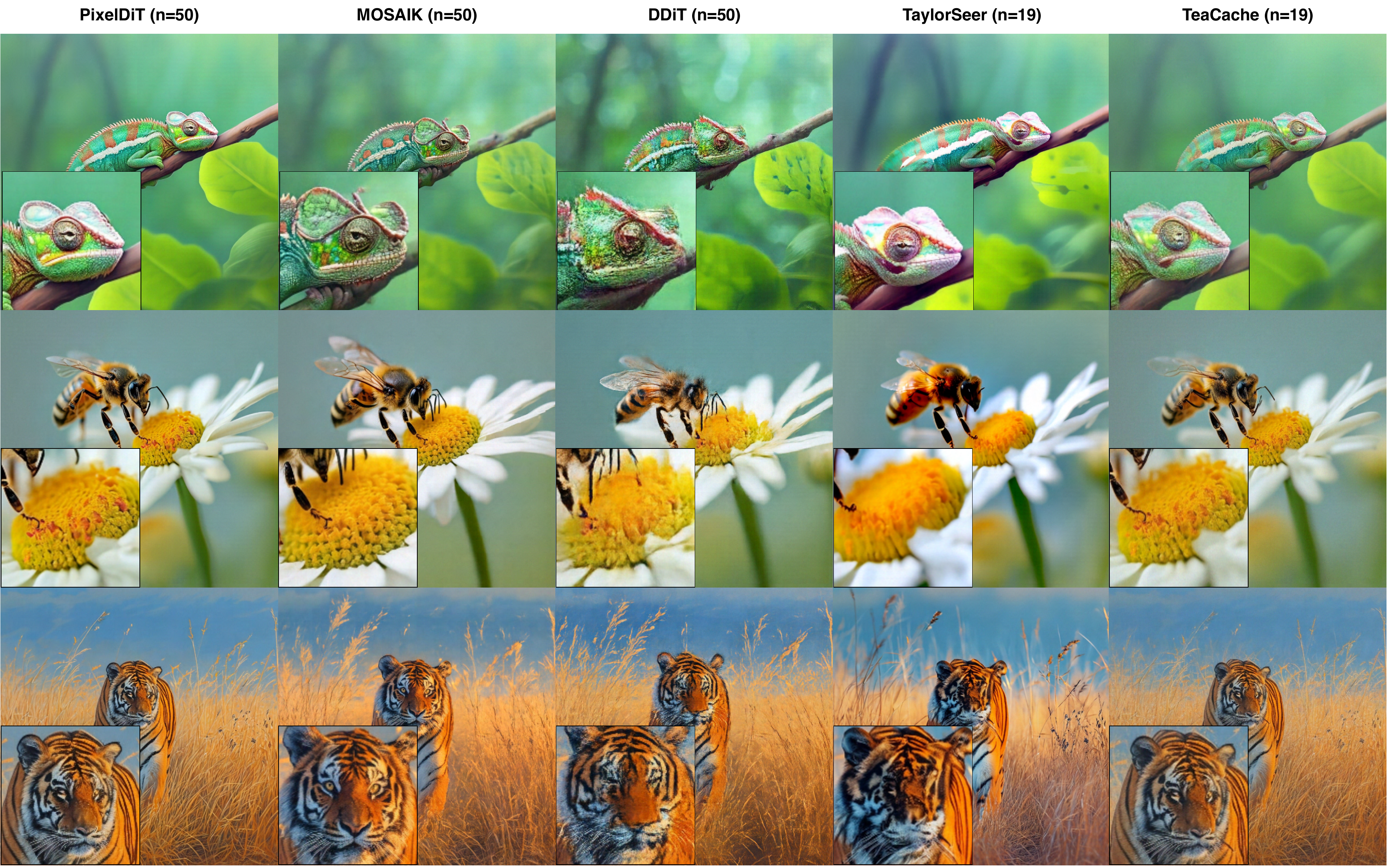}
\caption{\textbf{Qualitative comparisons} of \mosaik{} against DDiT, TeaCache, and TaylorSeer at a $60\%$ FLOPs reduction ($\sim$73\% token reduction). The base model (i.e., PixelDiT) with the highest FLOPs is included as a reference. Zoomed-in regions are provided to better illustrate fine-grained details.}
\label{fig:quality_comp}
\end{figure*}

During inference, only the first step uses a uniform $p16$ layout to produce the initial activations. Later, at each denoising step $t$, the predictor uses the activation from the previous timestep $\mathbf{h}_6^{t-1}$. The predictor assigns a damage score $d^t_r$ to each region $r$. We initialize all regions at $p64$ and consider two successive refinement actions. Refining a region from $p64$ to $p32$ adds three tokens and is assigned priority $d^t_r/3$, while refining it from $p32$ to $p16$ adds twelve tokens and is assigned priority $d^t_r/12$. At each iteration, we apply the highest-priority available refinement that fits within the remaining budget. We empirically found that $p64$ to $p16$ damage reliably guides both refinements across scales (see Appendix). Regions with high predicted damage are therefore more likely to reach $p16$, while regions with lower damage remain at coarser patch sizes. As illustrated in Fig.~\ref{fig:patch_layout_progress}, fine patches follow high-damage regions over the denoising trajectory, while low-damage regions remain coarse. Because training is budget-agnostic, the same model can operate at any inference budget, determining how many patches of each size to use and assigning finer patches where coarsening would cause the greatest degradation. 
\section{Results and Discussions}
\label{sec:results}

\subsection{Experimental Setup}

We adopt PixelDiT~\citep{pixeldit2025} at 1024px resolution as our pixel-space diffusion backbone, paired with a Gemma-2 text encoder~\citep{team2024gemma}. PixelDiT's architecture is extended for heterogeneous patch processing via LoRA with rank 32. Stages 1 and 2 are trained using AdamW~\citep{loshchilov2017decoupled} on BLIP3o-60k~\cite{chen2025blip3o} and text-to-image-2M-1024~\cite{zou2024text2image} with a batch size of 96 for 114{,}000 and 20{,}000 steps, respectively. The layout predictor is subsequently trained offline on intermediate features extracted from the Stage~2 multi-patch generator. Implementation and training details are provided in the Appendix.

\definecolor{avghl}{gray}{0.93}

\newcolumntype{M}{l<{\hspace{15pt}}} 
\newcolumntype{T}{c<{\hspace{15pt}}} 
\newcolumntype{O}{c<{\hspace{15pt}}} 

\begin{table*}[t]
\centering
\small
\setlength{\tabcolsep}{2.25pt}
\renewcommand{\arraystretch}{1.19}

\begin{tabular}{M T cccccO ccccccc}
\toprule
& &
\multicolumn{6}{c}{\textbf{DPG-Bench $\uparrow$}}
&
\multicolumn{7}{c}{\textbf{GenEval $\uparrow$}} \\
\cmidrule(lr){3-8}
\cmidrule(lr){9-15}

\textbf{Method}
& \textbf{TR\,(\%)}
& \cellcolor{avghl}\textbf{Avg}
& \textbf{Attr}
& \textbf{Ent}
& \textbf{Rel}
& \textbf{Glob}
& \textbf{Oth}
& \cellcolor{avghl}\textbf{Avg}
& \textbf{S.Obj}
& \textbf{T.Obj}
& \textbf{Cnt}
& \textbf{Col}
& \textbf{Pos}
& \textbf{C.Att} \\

\midrule

PixelDiT ($n{=}50$)
& -
& \cellcolor{avghl}83.5
& 87.8
& 88.6
& 91.2
& 83.0
& 89.6
& \cellcolor{avghl}\textbf{0.74}
& 1.00
& 0.95
& 0.55
& 0.88
& 0.41
& 0.68 \\

\midrule
\multicolumn{15}{l}{%
    \textit{Light: $\sim$20\% FLOPs reduction}%
} \\

PixelDiT ($n{=}38$)
& -
& \cellcolor{avghl}83.3
& 87.7
& 89.3
& 93.3
& \textbf{84.5}
& 70.8
& \cellcolor{avghl}0.70
& \underline{0.98}
& 0.88
& \underline{0.56}
& 0.84
& 0.40
& 0.53 \\

TeaCache
& -
& \cellcolor{avghl}83.5
& \underline{88.3}
& 89.3
& \underline{93.4}
& 83.6
& \underline{72.8}
& \cellcolor{avghl}\underline{0.72}
& \underline{0.98}
& \textbf{0.92}
& 0.54
& 0.86
& 0.40
& \textbf{0.62} \\

TaylorSeer
& -
& \cellcolor{avghl}\textbf{83.7}
& \textbf{88.5}
& \underline{89.7}
& \textbf{93.6}
& \underline{83.9}
& 72.4
& \cellcolor{avghl}0.71
& \underline{0.98}
& \underline{0.90}
& 0.53
& 0.84
& 0.40
& 0.58 \\

DDiT$^\dagger$
& 27
& \cellcolor{avghl}83.1
& 87.9
& 89.1
& 93.3
& 83.0
& \textbf{73.6}
& \cellcolor{avghl}0.69
& 0.97
& 0.85
& 0.51
& \underline{0.88}
& \underline{0.41}
& 0.55 \\

\rowcolor{gray!15}
\textsc{MOSAIK} (ours)
& 25
& \cellcolor{avghl}\underline{83.6}
& 87.9
& \textbf{89.7}
& 93.0
& \underline{83.9}
& \textbf{73.6}
& \cellcolor{avghl}\textbf{0.74}
& \textbf{0.99}
& 0.89
& \textbf{0.59}
& \textbf{0.89}
& \textbf{0.47}
& \underline{0.60} \\

\midrule
\multicolumn{15}{l}{%
    \textit{Moderate: $\sim$50\% FLOPs reduction}%
} \\

PixelDiT ($n{=}24$)
& -
& \cellcolor{avghl}82.5
& 87.9
& 88.4
& 92.6
& \underline{83.6}
& 68.8
& \cellcolor{avghl}0.67
& 0.98
& \underline{0.86}
& 0.51
& 0.81
& 0.39
& 0.48 \\

TeaCache
& -
& \cellcolor{avghl}\underline{83.5}
& \textbf{88.2}
& 89.3
& \underline{92.9}
& 83.3
& \underline{72.8}
& \cellcolor{avghl}\underline{0.72}
& \underline{0.99}
& \textbf{0.91}
& \underline{0.54}
& \underline{0.85}
& \underline{0.41}
& \underline{0.62} \\

TaylorSeer
& -
& \cellcolor{avghl}\textbf{83.6}
& \underline{88.0}
& \textbf{89.6}
& \textbf{93.3}
& \textbf{83.9}
& \textbf{73.2}
& \cellcolor{avghl}0.71
& 0.98
& \textbf{0.91}
& 0.53
& 0.84
& \underline{0.41}
& 0.61 \\

DDiT$^\dagger$
& 64
& \cellcolor{avghl}81.6
& 86.8
& 88.0
& 92.6
& 79.6
& 69.6
& \cellcolor{avghl}0.65
& 0.96
& 0.78
& 0.49
& 0.82
& 0.35
& 0.51 \\

\rowcolor{gray!15}
\textsc{MOSAIK} (ours)
& 59
& \cellcolor{avghl}83.4
& 87.8
& \underline{89.5}
& 92.8
& \textbf{83.9}
& 72.4
& \cellcolor{avghl}\textbf{0.74}
& \textbf{1.00}
& \textbf{0.91}
& \textbf{0.55}
& \textbf{0.88}
& \textbf{0.46}
& \textbf{0.64} \\

\midrule
\multicolumn{15}{l}{%
    \textit{Aggressive: $\sim$60\% FLOPs reduction}%
} \\

PixelDiT ($n{=}19$)
& -
& \cellcolor{avghl}81.5
& 87.2
& 87.9
& 92.5
& 83.0
& 67.2
& \cellcolor{avghl}0.65
& \underline{0.98}
& 0.84
& 0.52
& 0.81
& 0.34
& 0.43 \\

TeaCache
& -
& \cellcolor{avghl}\underline{82.8}
& \underline{87.7}
& 88.9
& \textbf{93.2}
& \underline{83.6}
& 71.2
& \cellcolor{avghl}\underline{0.71}
& \textbf{0.99}
& \textbf{0.88}
& \underline{0.55}
& \underline{0.84}
& \underline{0.41}
& \underline{0.60} \\

TaylorSeer
& -
& \cellcolor{avghl}82.7
& 87.1
& \underline{89.2}
& 93.0
& 82.4
& 68.4
& \cellcolor{avghl}0.68
& \underline{0.98}
& \underline{0.87}
& 0.50
& 0.80
& 0.40
& 0.51 \\

DDiT$^\dagger$
& 72
& \cellcolor{avghl}81.5
& 86.4
& 88.8
& \underline{93.2}
& 81.2
& \underline{71.6}
& \cellcolor{avghl}0.65
& 0.96
& 0.77
& 0.48
& 0.82
& 0.34
& 0.52 \\

\rowcolor{gray!15}
\textsc{MOSAIK} (ours)
& 74
& \cellcolor{avghl}\textbf{83.1}
& \textbf{87.7}
& \textbf{89.8}
& 93.0
& \textbf{84.2}
& \textbf{72.0}
& \cellcolor{avghl}\textbf{0.74}
& \textbf{0.99}
& \textbf{0.88}
& \textbf{0.58}
& \textbf{0.88}
& \textbf{0.47}
& \textbf{0.61} \\

\midrule
\multicolumn{15}{l}{%
    \textit{Maximal: $\sim$70\% FLOPs reduction}%
} \\

PixelDiT ($n{=}16$)
& -
& \cellcolor{avghl}80.6
& 86.2
& 87.2
& 92.5
& 81.2
& 67.2
& \cellcolor{avghl}0.64
& \underline{0.98}
& \underline{0.83}
& 0.47
& 0.81
& 0.35
& 0.40 \\

TeaCache
& -
& \cellcolor{avghl}\textbf{82.6}
& \textbf{87.6}
& 88.8
& \textbf{92.9}
& \underline{81.8}
& \underline{69.6}
& \cellcolor{avghl}\underline{0.71}
& \textbf{0.99}
& \textbf{0.88}
& \underline{0.54}
& \underline{0.84}
& \underline{0.41}
& \underline{0.60} \\

TaylorSeer
& -
& \cellcolor{avghl}82.4
& 86.8
& \underline{89.0}
& \underline{92.9}
& \underline{81.8}
& \underline{69.6}
& \cellcolor{avghl}0.68
& \underline{0.98}
& \textbf{0.88}
& 0.50
& 0.80
& 0.39
& 0.49 \\

DDiT$^\dagger$
& 84
& \cellcolor{avghl}80.8
& 86.4
& 88.2
& 92.4
& 80.2
& 70.4
& \cellcolor{avghl}0.64
& 0.95
& 0.76
& 0.44
& 0.83
& 0.36
& 0.47 \\

\rowcolor{gray!15}
\textsc{MOSAIK} (ours)
& 83
& \cellcolor{avghl}\underline{82.5}
& \underline{87.3}
& \textbf{89.1}
& 92.5
& \textbf{82.1}
& \textbf{72.0}
& \cellcolor{avghl}\textbf{0.74}
& \textbf{0.99}
& \textbf{0.88}
& \textbf{0.57}
& \textbf{0.88}
& \textbf{0.47}
& \textbf{0.62} \\

\bottomrule
\end{tabular}

\caption{\textbf{Quantitative comparison against step reduction, feature caching, and temporal patch scheduling.} TR\,(\%) is the token reduction relative to the patch-16 baseline ($4096$ tokens). \textbf{Bold} and \underline{underline} mark the best and second-best score per column within each compute tier. $\dagger$: Derived from schedule-averaged tokens.}
\label{tab:main}
\end{table*}

All images are generated at 1024px with a 50-step Euler sampler and classifier-free guidance~(CFG)~\citep{ho2022classifier} of $2.75$, with fixed random seeds so that every comparison starts from identical initial noise. We report quality on GenEval~\cite{ghosh2023geneval} (553 prompts, 4 samples each) and DPG-Bench~\cite{hu2024ella} (1065 prompts, 4 samples each). As the official DDiT code is not public, we replicate its method on PixelDiT (marked $\dagger$), applying its temporal patch schedule with the proposed third-order finite-difference heuristic and thresholds calibrated to our compute tiers. We utilize the stage 1 checkpoint for DDiT, as it is optimized for uniform generation. 

To closely follow the evaluation protocol of our primary baseline (DDiT), we also include TeaCache~\cite{liu2025timestep} and TaylorSeer~\cite{liu2025reusing} as representative caching-based efficiency baselines. Efficiency is measured by the mean tokens per sampling step and relative FLOPs, defined as total image FLOPs as a fraction of the uniform $p16$, 50-step baseline, with per-step FLOPs averaged over the trajectory.

\subsection{Main Results}
\label{sec:main_results}

\begin{figure}[t] 
    \centering
    \includegraphics[width=0.85\linewidth]{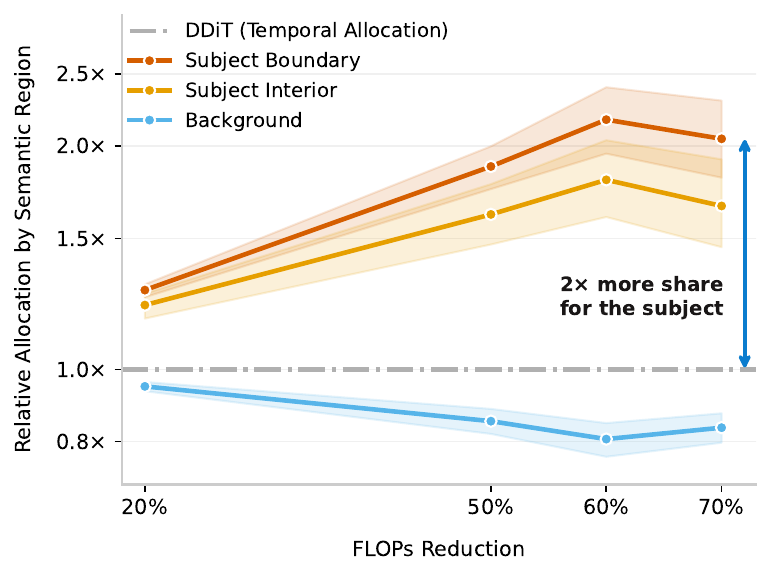}
    \caption{\textbf{Subject Compute Allocation vs. Budget.} Token allocation is measured against ground-truth semantic reference masks.}
    \label{fig:allocation_main}
\end{figure}

In Table~\ref{tab:main}, we present a quantitative comparison with the baseline and competing methods and group all methods into tiers of matched FLOPs reduction, so every ranking is read at approximately equal cost. The baseline achieves $0.74$ on GenEval and $83.5$ on DPG with $4096$ tokens per step. \mosaik{} reaches $0.74$ on GenEval and $82.5$ on DPG with $683$ tokens per step, corresponding to an $83\%$ token reduction and a $70\%$ reduction in FLOPs. Despite this aggressive token reduction, \mosaik{} matches the baseline on GenEval and remains only $1.0$ point below it on DPG. This supports our claim that uniform fine patchification spends unnecessary computation in regions that are less sensitive to coarsening. \mosaik{} also exhibits strong stability across compute budgets. Spanning a $25\%$ to $83\%$ token reduction, GenEval stays close to $0.74$ and DPG varies by only one point, while all other methods steadily degrade. By prioritizing low-damage regions for coarsening, the layout predictor preserves quality until tighter budgets force larger patches in sensitive regions~(see Fig.~\ref{fig:allocation_main}). In contrast, other methods coarsen all regions uniformly, causing fidelity to decline simultaneously in both foreground subjects and background areas.

This contrast is visually evident in our qualitative comparisons (Fig.~\ref{fig:quality_comp}). Under a similar FLOPs budget, whether by reducing the token count by $73\%$ (for DDiT and \mosaik) or setting $n\!=\!19$ sampling steps~(for TaylorSeer and TeaCache), \mosaik{} preserves a level of fine-grained, high-frequency detail comparable to that of the full-compute PixelDiT baseline.
In contrast, DDiT suffers from localized blurring and artifacts. As seen in the zoomed-in regions, it tends to over-smooth high-frequency areas and lose granular details such as flower pollen and tiger textures. This demonstrates the clear advantage of \mosaik{} to effectively avoid such degradation and maintain structural integrity.

\paragraph{Better than temporal patch scheduling.}
In DDiT, patch size cannot vary across spatial regions within the same forward pass. When DDiT detects rapid change along the denoising trajectory, it switches the entire layout to a finer patch size, allocating additional tokens even to smooth background regions. Once the trajectory stabilizes, it applies a coarser patch size globally, which can remove important details from the primary subject. 
As detailed in Fig.~\ref{fig:allocation_main}, DDiT's temporal scheduling locks its relative compute allocation at exactly $1.0\times$ across all semantic regions. In contrast, \mosaik{} actively redirects compute based on demand. As the FLOPs reduction scales to $60\%$, \mosaik{} progressively starves the background (dropping to $0.8\times$ relative allocation) to route tokens to the subject. 
Crucially, it dynamically distinguishes within the subject itself, prioritizing the highly complex subject boundary over the subject interior. 

Table~\ref{tab:main} also shows that, at a modest $27\%$ token reduction, DDiT already falls $0.05$ GenEval points below \mosaik{}, and at its most aggressive setting, DDiT's performance collapses to $0.64$, while \mosaik{} maintains a score of $0.74$. The DPG metric exhibits a similar trend. These results demonstrate that a purely temporal schedule cannot adequately capture the spatial heterogeneity of representational demand.

\paragraph{Better than fewer denoising steps.}
Reducing PixelDiT steps matches our FLOPs reductions, but quality falls monotonically to $0.64$ GenEval and $80.6$ DPG at $70\%$: truncating the denoising trajectory removes the late refinement steps required across the image. In contrast, \mosaik{} retains all 50 steps while reducing the total cost, and leads at $70\%$ by great margin of $0.1$ on GenEval and $1.9$ on DPG.

\begin{table}[t]
\centering
\small
\setlength{\tabcolsep}{4pt}
\renewcommand{\arraystretch}{0.9}
\begin{tabular}{l c c c}
\toprule
\textbf{Method} & \textbf{TR\,(\%)} & \textbf{DPG-Bench} & \textbf{GenEval} \\

\specialrule{0.10em}{0.3em}{0.3em}
\multicolumn{4}{l}{\textit{$\sim$70\% FLOPs reduction}} \\
TeaCache    & - & 82.6 & 0.71 \\
\rowcolor{gray!15} \textsc{MOSAIK} $+$ TeaCache & 73 & \textbf{82.8} & \textbf{0.74} \\

\cmidrule[0.015em](lr){1-4}

TaylorSeer  & - & 82.4 & 0.68 \\
\rowcolor{gray!15} \textsc{MOSAIK} $+$ TaylorSeer & 73 & \textbf{82.8} & \textbf{0.74} \\

\specialrule{0.10em}{0.3em}{0.3em}
\multicolumn{4}{l}{\textit{$\sim$80\% FLOPs reduction}} \\
TeaCache    & - & 80.7 & 0.67 \\
\rowcolor{gray!15} \textsc{MOSAIK} $+$ TeaCache & 83 & \textbf{82.2} & \textbf{0.73} \\

\cmidrule[0.015em](lr){1-4}

TaylorSeer  & - & 72.9 & 0.43 \\
\rowcolor{gray!15} \textsc{MOSAIK} $+$ TaylorSeer & 83 & \textbf{82.2} & \textbf{0.73} \\
\bottomrule
\end{tabular}
\caption{\textbf{Composing the spatial and temporal axes.} \textsc{MOSAIK} integrates seamlessly with caching to enhance performance under aggressive compression.}
\label{tab:caching}
\end{table}

\paragraph{Complementary to feature caching.}
TeaCache and TaylorSeer reuse activations across denoising steps, whereas \mosaik{} spatially coarsens regions within each step. Because caching preserves quality better than step truncation, these methods provide strong baselines at matched computational budgets (Table~\ref{tab:main}). Even against these baselines, \mosaik{} consistently achieves the highest GenEval scores, with its margin over TeaCache widening under tighter compute constraints. This advantage stems from their differing approximations: caching skips temporal computation, whereas \mosaik{} performs a cheaper forward pass at every step, reducing computation only in insensitive regions. Consequently, the mechanisms are complementary, naturally combining temporal efficiency across steps with spatial efficiency through adaptive token allocation. Table~\ref{tab:caching} shows what the composition buys: when we combine with caching, we can use more tokens while maintaining low FLOPs. At a matched budget, adding \mosaik{} on top of either caching method improves both benchmarks over that method alone, and the gain grows precisely where caching is weakest: at an $80\%$ reduction, caching alone degrades sharply and TaylorSeer collapses, while the composed arms hold near-headline GenEval and restore DPG. The composition therefore extends the compute-quality frontier in both directions: at moderate budgets it recovers the composition caching loses, and at aggressive budgets it sustains quality at reductions neither axis reaches by itself.

\newcommand{\gain}[1]{{\footnotesize$+#1$}}
\begin{table}[t]
  \centering
  \setlength{\tabcolsep}{4pt}
  \renewcommand{\arraystretch}{0.9}
  \footnotesize
  \begin{tabular}{ll cccc}
    \toprule
    \multirow{2}{*}{Benchmark} & \multirow{2}{*}{Patch layout}
      & \multicolumn{4}{c}{FLOPs reduction (\%)} \\
    \cmidrule(lr){3-6}
      & & $20$ & $50$ & $60$ & $70$ \\
    \midrule
    \multirow{2}{*}{DPG}
      & Random           & 83.0 & 82.1 & 82.1 & 81.6 \\
      & Damage-guided    & \textbf{83.6} & \textbf{83.4} & \textbf{83.1} & \textbf{82.5} \\
    \midrule
    \multirow{2}{*}{GenEval}
      & Random           & 0.72 & 0.72 & 0.71 & 0.70 \\
      & Damage-guided    & \textbf{0.74} & \textbf{0.74} & \textbf{0.74} & \textbf{0.74} \\
    \bottomrule
  \end{tabular}
  \caption{\textbf{The gain is placement.} The same multi-patch generator at an identical token budget,
  driven by a random layout vs.\ the damage-guided predictor.}
  \label{tab:ablation_blind}
\end{table}
\paragraph{Efficacy of the damage-guided layout predictor.}
\label{sec:ablation}
Our Stage~2 enables multi-patch generation, but it is the damage-guided layout predictor that unlocks the efficiency gains driving our core motivation. Under a fixed multi-patch generator, computational budget, and initialization seed, substituting our damage-guided layout predictor with a content-blind baseline results in both GenEval and DPG-bench degradation across budgets (Table~\ref{tab:ablation_blind}). This performance gap widens under tighter computational constraints, demonstrating the efficacy of the damage predictor, which preserves quality by prioritizing finer patches in highly sensitive regions. We validate the causal role of the damage predictor and its near-optimality compared to ground-truth measurements in Appendix.
\section{Conclusion}
\label{sec:conclusion}
We introduce \textsc{MOSAIK}, a damage-guided framework that dynamically varies patch size across image regions and denoising steps. A lightweight predictor estimates the fidelity loss
caused by coarsening, enabling \textsc{MOSAIK} to allocate fine patches where they are most needed.
Consequently, it matches full-compute PixelDiT fidelity while substantially reducing 
the token count and compute overhead. Compared to temporal patch scheduling techniques, \textsc{MOSAIK} consistently outperforms temporal patch scheduling,
with its advantage widening under aggressive budget cuts. Moreover, it can be combined with caching techniques to achieve further FLOPs reduction while maintaining quality. Future work will extend this to video diffusion, where redundancy can be exploited jointly across space and time.

\clearpage
\bibliography{aaai2027}

@article{hu2022lora,
  title={Lora: Low-rank adaptation of large language models.},
  author={Hu, Edward J and Shen, Yelong and Wallis, Phillip and Allen-Zhu, Zeyuan and Li, Yuanzhi and Wang, Shean and Wang, Liang and Chen, Weizhu and others},
  journal={Iclr},
  volume={1},
  number={2},
  pages={3},
  year={2022}
}

@article{team2024gemma,
  title={Gemma 2: Improving open language models at a practical size},
  author={Team, Gemma and Riviere, Morgane and Pathak, Shreya and Sessa, Pier Giuseppe and Hardin, Cassidy and Bhupatiraju, Surya and Hussenot, L{\'e}onard and Mesnard, Thomas and Shahriari, Bobak and Ram{\'e}, Alexandre and others},
  journal={arXiv preprint arXiv:2408.00118},
  year={2024}
}

@inproceedings{ldm2022,
  title     = {High-Resolution Image Synthesis With Latent Diffusion Models},
  author    = {Rombach, Robin and Blattmann, Andreas and Lorenz, Dominik and Esser, Patrick and Ommer, Bj{\"o}rn},
  booktitle = {Proceedings of the IEEE/CVF Conference on Computer Vision and Pattern Recognition (CVPR)},
  month     = {June},
  year      = {2022},
  pages     = {10684--10695}
}

@inproceedings{dit2023,
  title     = {Scalable Diffusion Models with Transformers},
  author    = {Peebles, William and Xie, Saining},
  booktitle = {Proceedings of the IEEE/CVF International Conference on Computer Vision (ICCV)},
  month     = {October},
  year      = {2023},
  pages     = {4195--4205}
}

@inproceedings{dcae2024,
  title     = {Deep Compression Autoencoder for Efficient High-Resolution Diffusion Models},
  author    = {Chen, Junyu and Cai, Han and Chen, Junsong and Xie, Enze and Yang, Shang and Tang, Haotian and Li, Muyang and Han, Song},
  booktitle = {International Conference on Learning Representations (ICLR)},
  year      = {2025}
}

@inproceedings{jit2025,
  title     = {Back to Basics: Let Denoising Generative Models Denoise},
  author    = {Li, Tianhong and He, Kaiming},
  booktitle = {Proceedings of the IEEE/CVF Conference on Computer Vision and Pattern Recognition (CVPR)},
  month     = {June},
  year      = {2026},
  pages     = {36115--36125}
}

@article{deco2025,
  title   = {{DeCo}: Frequency-Decoupled Pixel Diffusion for End-to-End Image Generation},
  author  = {Ma, Zehong and Wei, Longhui and Wang, Shuai and Zhang, Shiliang and Tian, Qi},
  journal = {arXiv preprint arXiv:2511.19365},
  year    = {2025}
}

@article{dip2025,
  title   = {{DiP}: Taming Diffusion Models in Pixel Space},
  author  = {Chen, Zhennan and Zhu, Junwei and Chen, Xu and Zhang, Jiangning and Hu, Xiaobin and Zhao, Hanzhen and Wang, Chengjie and Yang, Jian and Tai, Ying},
  journal = {arXiv preprint arXiv:2511.18822},
  year    = {2025}
}

@inproceedings{pixeldit2025,
  title     = {{PixelDiT}: Pixel Diffusion Transformers for Image Generation},
  author    = {Yu, Yongsheng and Xiong, Wei and Nie, Weili and Sheng, Yichen and Liu, Shiqiu and Luo, Jiebo},
  booktitle = {Proceedings of the IEEE/CVF Conference on Computer Vision and Pattern Recognition (CVPR)},
  month     = {June},
  year      = {2026},
  pages     = {14273--14282}
}

@article{hyperdit2026,
  title   = {{HyperDiT}: Hyper-Connected Transformers for High-Fidelity Pixel-Space Diffusion},
  author  = {He, Yu and Ma, Lichen and Guo, Zipeng and Shan, Xinyuan and Fu, Jingling and Chen, Dong and Huang, Junshi and Li, Yan},
  journal = {arXiv preprint arXiv:2605.15741},
  year    = {2026}
}

@inproceedings{ddit2026,
  title     = {{DDiT}: Dynamic Patch Scheduling for Efficient Diffusion Transformers},
  author    = {Kim, Dahye and Ghadiyaram, Deepti and Gadde, Raghudeep},
  booktitle = {Proceedings of the IEEE/CVF Conference on Computer Vision and Pattern Recognition (CVPR)},
  month     = {June},
  year      = {2026},
  pages     = {11459--11471}
}

@article{ppflow2026,
  title   = {Pyramidal Patchification Flow for Visual Generation},
  author  = {Li, Hui and Chen, Baoyou and Zhang, Liwei and Li, Jiaye and Wang, Jingdong and Zhu, Siyu},
  journal = {arXiv preprint arXiv:2506.23543},
  year    = {2025}
}

@inproceedings{mpdit2026,
  title     = {{MPDiT}: Multi-Patch Global-to-Local Transformer Architecture For Efficient Flow Matching and Diffusion Model},
  author    = {Dao, Quan and Metaxas, Dimitris},
  booktitle = {Proceedings of the IEEE/CVF Conference on Computer Vision and Pattern Recognition (CVPR)},
  month     = {June},
  year      = {2026},
  pages     = {33000--33011}
}

@inproceedings{tome2023,
  title     = {Token Merging: Your {ViT} But Faster},
  author    = {Bolya, Daniel and Fu, Cheng-Yang and Dai, Xiaoliang and Zhang, Peizhao and Feichtenhofer, Christoph and Hoffman, Judy},
  booktitle = {International Conference on Learning Representations (ICLR)},
  year      = {2023}
}

@inproceedings{atedm2024,
  title     = {Attention-Driven Training-Free Efficiency Enhancement of Diffusion Models},
  author    = {Wang, Hongjie and Liu, Difan and Kang, Yan and Li, Yijun and Lin, Zhe and Jha, Niraj K. and Liu, Yuchen},
  booktitle = {Proceedings of the IEEE/CVF Conference on Computer Vision and Pattern Recognition (CVPR)},
  month     = {June},
  year      = {2024},
  pages     = {16080--16089}
}

@inproceedings{dydit2025,
  title     = {Dynamic Diffusion Transformer},
  author    = {Zhao, Wangbo and Han, Yizeng and Tang, Jiasheng and Wang, Kai and Song, Yibing and Huang, Gao and Wang, Fan and You, Yang},
  booktitle = {International Conference on Learning Representations (ICLR)},
  year      = {2025}
}

@inproceedings{diffcr2025,
  title     = {Layer- and Timestep-Adaptive Differentiable Token Compression Ratios for Efficient Diffusion Transformers},
  author    = {You, Haoran and Barnes, Connelly and Zhou, Yuqian and Kang, Yan and Du, Zhenbang and Zhou, Wei and Zhang, Lingzhi and Nitzan, Yotam and Liu, Xiaoyang and Lin, Zhe and Shechtman, Eli and Amirghodsi, Sohrab and Lin, Yingyan Celine},
  booktitle = {Proceedings of the IEEE/CVF Conference on Computer Vision and Pattern Recognition (CVPR)},
  month     = {June},
  year      = {2025},
  pages     = {18072--18082}
}

@inproceedings{msvit2023,
  title     = {{MSViT}: Dynamic Mixed-Scale Tokenization for Vision Transformers},
  author    = {Havtorn, Jakob Drachmann and Royer, Am{\'e}lie and Blankevoort, Tijmen and Bejnordi, Babak Ehteshami},
  booktitle = {Proceedings of the IEEE/CVF International Conference on Computer Vision (ICCV) Workshops},
  month     = {October},
  year      = {2023},
  pages     = {838--848}
}

@article{apt2026,
  title   = {Accelerating Vision Transformers with Adaptive Patch Sizes},
  author  = {Choudhury, Rohan and Kim, JungEun and Park, Jinhyung and Yang, Eunho and Jeni, L{\'a}szl{\'o} A. and Kitani, Kris M.},
  journal = {arXiv preprint arXiv:2510.18091},
  year    = {2025}
}

@inproceedings{d2it2025,
  title     = {{D$^2$iT}: Dynamic Diffusion Transformer for Accurate Image Generation},
  author    = {Jia, Weinan and Huang, Mengqi and Chen, Nan and Zhang, Lei and Mao, Zhendong},
  booktitle = {Proceedings of the IEEE/CVF Conference on Computer Vision and Pattern Recognition (CVPR)},
  month     = {June},
  year      = {2025},
  pages     = {12860--12870}
}

@article{foveated2026,
  title   = {Foveated Diffusion: Efficient Spatially Adaptive Image and Video Generation},
  author  = {Chao, Brian and Yariv, Lior and Xiao, Howard and Wetzstein, Gordon},
  journal = {arXiv preprint arXiv:2603.23491},
  year    = {2026}
}

@inproceedings{patchscaler2025,
  title     = {{PatchScaler}: An Efficient Patch-Independent Diffusion Model for Image Super-Resolution},
  author    = {Liu, Yong and Dong, Hang and Pan, Jinshan and Dong, Qingji and Chen, Kai and Zhang, Rongxiang and Fu, Lean and Wang, Fei},
  booktitle = {Proceedings of the IEEE/CVF International Conference on Computer Vision (ICCV)},
  month     = {October},
  year      = {2025},
  pages     = {11283--11293}
}

@inproceedings{dynapatch2026,
  title     = {Content-Aware Dynamic Patchification for Efficient Video Diffusion},
  author    = {Li, Sheng and Barnes, Connelly and Rizve, Mamshad Nayeem and Peng, Hongwu and Li, Zhengang and Dibua, Ohi and Ganjdanesh, Alireza and Tang, Xulong and Kang, Yan and Gong, Yifan},
  booktitle = {Proceedings of the IEEE/CVF Conference on Computer Vision and Pattern Recognition (CVPR)},
  month     = {June},
  year      = {2026},
  pages     = {35936--35945}
}

@inproceedings{adaptok2026,
  title     = {Learning Adaptive and Temporally Causal Video Tokenization in a 1D Latent Space},
  author    = {Li, Yan and Tian, Changyao and Xia, Renqiu and Liao, Ning and Guo, Weiwei and Yan, Junchi and Li, Hongsheng and Dai, Jifeng and Li, Hao and Yang, Xue},
  booktitle = {Proceedings of the IEEE/CVF Conference on Computer Vision and Pattern Recognition (CVPR)},
  year      = {2026}
}

@inproceedings{ho2020ddpm,
  title     = {Denoising Diffusion Probabilistic Models},
  author    = {Ho, Jonathan and Jain, Ajay and Abbeel, Pieter},
  booktitle = {Advances in Neural Information Processing Systems (NeurIPS)},
  year      = {2020}
}

@inproceedings{lipman2023flow,
  title     = {Flow Matching for Generative Modeling},
  author    = {Lipman, Yaron and Chen, Ricky T. Q. and Ben-Hamu, Heli and Nickel, Maximilian and Le, Matt},
  booktitle = {International Conference on Learning Representations (ICLR)},
  year      = {2023}
}

@article{su2024roformer,
  title   = {{RoFormer}: Enhanced Transformer with Rotary Position Embedding},
  author  = {Su, Jianlin and Lu, Yu and Pan, Shengfeng and Murtadha, Ahmed and Wen, Bo and Liu, Yunfeng},
  journal = {Neurocomputing},
  volume  = {568},
  pages   = {127063},
  year    = {2024}
}

@article{flexidit2025,
  title   = {{FlexiDiT}: Your Diffusion Transformer Can Easily Generate High-Quality Samples with Less Compute},
  author  = {Anagnostidis, Sotiris and Bachmann, Gregor and Kim, Yeongmin and Kohler, Jonas and Georgopoulos, Markos and Sanakoyeu, Artsiom and Du, Yuming and Pumarola, Albert and Thabet, Ali and Sch{\"o}nfeld, Edgar},
  journal = {arXiv preprint arXiv:2502.20126},
  year    = {2025}
}

@inproceedings{vaswani2017attention,
  title     = {Attention Is All You Need},
  author    = {Vaswani, Ashish and Shazeer, Noam and Parmar, Niki and Uszkoreit, Jakob and Jones, Llion and Gomez, Aidan N. and Kaiser, Lukasz and Polosukhin, Illia},
  booktitle = {Advances in Neural Information Processing Systems (NeurIPS)},
  year      = {2017}
}

@inproceedings{repa2024,
  title     = {Representation Alignment for Generation: Training Diffusion Transformers Is Easier Than You Think},
  author    = {Yu, Sihyun and Kwak, Sangkyung and Jang, Huiwon and Jeong, Jongheon and Huang, Jonathan and Shin, Jinwoo and Xie, Saining},
  booktitle = {International Conference on Learning Representations (ICLR)},
  year      = {2025}
}

@inproceedings{perez2018film,
  title={{FiLM}: Visual Reasoning with a General Conditioning Layer},
  author={Perez, Ethan and Strub, Florian and de Vries, Harm and Dumoulin, Vincent and Courville, Aaron},
  booktitle={Proceedings of the AAAI Conference on Artificial Intelligence},
  year={2018}
}

@InProceedings{kingma2014auto,
title={Auto-Encoding Variational Bayes},
author={Kingma, Diederik P and Welling, Max},
booktitle={ICLR},
year={2014}
}

@article{rezende2014stochastic,
  title={Stochastic backpropagation and approximate inference in deep generative models},
  author={Rezende, Danilo Jimenez and Mohamed, Shakir and Wierstra, Daan},
  journal={ICML},
  year={2014}
}

@misc{chen2025blip3o,
      title={BLIP3-o: A Family of Fully Open Unified Multimodal Models-Architecture, Training and Dataset}, 
      author={Jiuhai Chen and Zhiyang Xu and Xichen Pan and Yushi Hu and Can Qin and Tom Goldstein and Lifu Huang and Tianyi Zhou and Saining Xie and Silvio Savarese and Le Xue and Caiming Xiong and Ran Xu},
      year={2025},
      eprint={2505.09568},
      archivePrefix={arXiv},
      primaryClass={cs.CV},
      url={https://arxiv.org/abs/2505.09568}
}

@misc{zou2024text2image,
      title={text-to-image-2M: A high-quality, diverse text--image training dataset},
      author={Kai Zou},
      year={2024},
      doi={10.57967/hf/3066}
}

@article{loshchilov2017decoupled,
  title={Decoupled weight decay regularization},
  author={Loshchilov, Ilya and Hutter, Frank},
  journal={arXiv preprint arXiv:1711.05101},
  year={2017}
}

@article{hu2024ella,
  title={Ella: Equip diffusion models with llm for enhanced semantic alignment},
  author={Hu, Xiwei and Wang, Rui and Fang, Yixiao and Fu, Bin and Cheng, Pei and Yu, Gang},
  journal={arXiv preprint arXiv:2403.05135},
  year={2024}
}

@article{ghosh2023geneval,
  title={Geneval: An object-focused framework for evaluating text-to-image alignment},
  author={Ghosh, Dhruba and Hajishirzi, Hannaneh and Schmidt, Ludwig},
  journal={Advances in Neural Information Processing Systems},
  volume={36},
  pages={52132--52152},
  year={2023}
}

@article{ho2022classifier,
  title={Classifier-free diffusion guidance},
  author={Ho, Jonathan and Salimans, Tim},
  journal={arXiv preprint arXiv:2207.12598},
  year={2022}
}

@inproceedings{liu2025timestep,
  title={Timestep Embedding Tells: It's Time to Cache for Video Diffusion Model},
  author={Liu, Feng and Zhang, Shiwei and Wang, Xiaofeng and Wei, Yujie and Qiu, Haonan and Zhao, Yuzhong and Zhang, Yingya and Ye, Qixiang and Wan, Fang},
  booktitle={Proceedings of the Computer Vision and Pattern Recognition Conference},
  pages={7353--7363},
  year={2025}
}

@inproceedings{liu2025reusing,
  title={From reusing to forecasting: Accelerating diffusion models with taylorseers},
  author={Liu, Jiacheng and Zou, Chang and Lyu, Yuanhuiyi and Chen, Junjie and Zhang, Linfeng},
  booktitle={Proceedings of the IEEE/CVF International Conference on Computer Vision},
  pages={15853--15863},
  year={2025}
}

\clearpage
\appendix 
\providecommand{\airev}[1]{\textcolor{airevcolor}{#1}}
\definecolor{airevcolor}{HTML}{1A5FB4}  

\section{Implementation and Training Details}
\label{sec:app_impl}

\subsection{Experimental Setup for Stage~1 \& Stage~2}
\label{sec:app_stages_details}

\paragraph{Architecture and Initialization.}
LoRA is applied to the attention, MLP, and adaptive-normalization layers of the DiT blocks and to the attention, MLP, and conditioning projections of the PiT decoder. The $p32$ and $p64$ tokenizers are initialized by composing the pretrained $p16$ projection with bilinear downsampling. Their pixel-shuffle upsamplers are identity initialized. Patch size embeddings are learned for $p32$ and $p64$, while $p16$ retains the original path. All pretrained PixelDiT weights stay frozen. The self-distillation teacher is that same frozen network with the adapters switched off and the multi-patch generator the same network with them enabled, so no second set of weights is stored.

\paragraph{Multi-Patch Generator Training.}
In Stage~1, each optimization step samples one uniform patch size for the entire minibatch, with probabilities $0.2$, $0.4$, and $0.4$ for $p16$, $p32$, and $p64$, respectively. We use a constant learning rate of $2\times10^{-4}$ and initialize Stage~2 from the checkpoint at $114{,}000$ optimization steps. For the mixed-layout branch of Stage~2, we sample the relative proportions of $p16$, $p32$, and $p64$ regions independently for each image, then assign each region a patch size according to those proportions. This spans a broad range of token budgets, and drawing the proportions without reference to the image leaves no spatial arrangement the generator could specialize to. The uniform layout branch samples one of the three patch sizes uniformly. Stage~2 uses a smaller learning rate of $10^{-4}$, and we select the checkpoint at $20{,}000$ optimization steps.

\subsection{Stage~3: Damage-Guided Layout Prediction}
\label{sec:app_stages_details_extended}

\paragraph{Mixed-Layout Trajectory Collection.}
We collect one $50$-step trajectory for each of $4{,}000$ prompts randomly sampled from the caption pool used in Stage~2, which combines BLIP3o-60k with a 10K native-$1024$ subset of text-to-image-2M. For each trajectory, we sample a heterogeneous layout using the procedure described in Section~\ref{sec:app_stages_details}, hold it fixed across denoising steps, and resample it for the next prompt. We retain eight approximately evenly spaced states after the initial step, producing $32{,}000$ records. At each retained state $t\geq1$, the target in
Eq.~2
is computed by evaluating the same noisy state $x_t$ with the uniform $p16$ teacher and the uniform $p64$ generator. The target is therefore a property of $x_t$ alone and cannot be biased by the mixed patch layout. The predictor input is the block-6 activation $\mathbf{h}_{6}^{t-1}$ from the preceding mixed-patch forward pass. Because its tokens cover different patch sizes, we reconstruct a dense feature map by repeating each token over its support on the $p16$ grid. We retain activations from the conditional classifier-free guidance (CFG) branch, matching the inference procedure.

\paragraph{Damage Predictor Architecture and Training.}
Using a $1\times1$ projection, $\mathbf{h}_{6}^{t-1}$ is mapped to a hidden dimension of $192$. Two convolutional layers capture local spatial structure, while one self-attention layer models interactions between distant regions. FiLM modulation conditions the intermediate features on the noise level $\sigma$ and pooled text embedding. The layout predictor contains approximately $1.63$M parameters and is trained offline using AdamW with a constant learning rate of $2\times10^{-3}$, a batch size of $256$, and gradient clipping at $1.0$. We validate every $500$ optimization steps and select the checkpoint with the highest Spearman rank correlation between predicted and measured regional damage. Rank correlation, not absolute regression error, is what determines behavior: the allocator reads the damage map only through the ordering of candidate refinements by score per added token, which is unchanged by a global rescaling. Because all inputs and targets are precomputed, predictor training does not require the diffusion backbone or text encoder.

\begin{figure}[t]
\centering
\includegraphics[width=\linewidth]{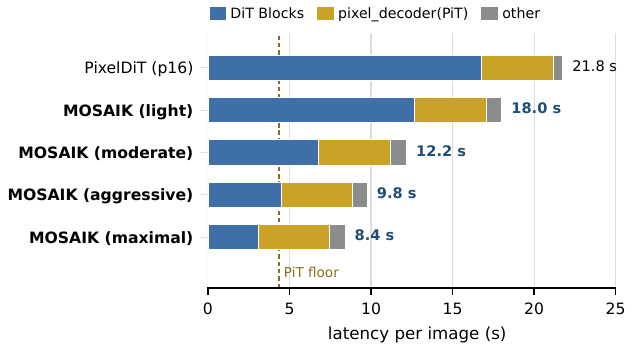}
\caption{\textbf{Measured per-image latency on consumer level GPU.} Speedups are relative to PixelDiT at uniform $p16$.}
\label{fig:app_latency}
\end{figure}

\begin{table}[t]
\centering
\footnotesize
\setlength{\tabcolsep}{3pt}
\renewcommand{\arraystretch}{1.1}
\begin{tabular}{@{}lrrrr@{}}
\toprule
\multirow{2}{*}{\textbf{Module}} & \multicolumn{3}{c}{\textbf{GFLOPs}} & \multirow{2}{*}{\textbf{Max $\downarrow$}} \\
\cmidrule(lr){2-4}
& \textbf{p16} & \textbf{p32} & \textbf{p64} & \\
\midrule
\multicolumn{5}{@{}l}{\textbf{Dynamic Compute (Scales with Patch)}} \\
\midrule
\texttt{DiT\_mlp\_img} & 2227 & 557 & 139 & 33\% \\
\texttt{DiT\_attn}     & 1662 & 151 & 27  & 26\% \\
\texttt{DiT\_qkv\_img} & 834  & 209 & 52  & 12\% \\
\texttt{DiT\_proj\_img}& 282  & 70  & 18  & 4\% \\
\midrule
\multicolumn{5}{@{}l}{\textbf{Fixed Overhead (Independent)}} \\
\midrule
\texttt{pixel\_decoder} (PiT) & 1068 & 1068 & 1068 & 0\% \\
\texttt{text\_stream}         & 245  & 245  & 245  & 0\% \\
Token/Head/Embed/adaLN        & 13   & 13   & 13   & 0\% \\
Upsampler                     & 0    & 19   & 19   & 0\% \\
\bottomrule
\end{tabular}
\caption{\textbf{Per-module GFLOPs breakdown across patch sizes.} \textbf{Max $\downarrow$} denotes each dynamic module's contribution to the FLOPs reduction from $p16$ to $p64$.}
\label{tab:flops_anatomy}
\end{table}

\begin{table*}[t]
\centering
\footnotesize
\setlength{\tabcolsep}{6pt}
\renewcommand{\arraystretch}{1.15}
\begin{tabular}{@{}lcccccccc@{}}
\toprule
\textbf{Configuration} & \textbf{TR\,(\%)} & \textbf{GenEval} & \textbf{S.Obj} & \textbf{T.Obj} & \textbf{Cnt} & \textbf{Pos} & \textbf{C.Att} & \textbf{DPG} \\
\midrule
\multicolumn{9}{@{}l}{\textit{Baseline (Anchor)}} \\
Stage~2 (Uniform $p16$) & $0$ & $0.73$ & $0.99$ & $0.90$ & $0.55$ & $0.47$ & $0.59$ & $83.4$ \\
\midrule
\multicolumn{9}{@{}l}{\textit{Aggressive: $\sim$60\% FLOPs reduction}} \\
Stage~2 (Uniform $p32$) & $75$ & $0.73$ & $1.00$ & $0.92$ & $0.55$ & $0.45$ & $0.59$ & $\mathbf{83.2}$ \\
\rowcolor{gray!15}
\mosaik{} (ours) & $74$ & $\mathbf{0.74}$ & $0.99$ & $0.88$ & $0.58$ & $0.47$ & $0.61$ & $83.1$ \\
\midrule
\multicolumn{9}{@{}l}{\textit{Maximal: $\sim$70\% FLOPs reduction}} \\
Stage~2 (Uniform $p64$) & $94$ & $0.65$ & $0.97$ & $0.78$ & $0.47$ & $0.42$ & $0.49$ & $80.2$ \\
\rowcolor{gray!15}
\mosaik{}~(ours) & $83$ & $\mathbf{0.74}$ & $0.99$ & $0.88$ & $0.57$ & $0.48$ & $0.63$ & $82.5$ \\
\rowcolor{gray!15}
\mosaik{}~(ours) + TeaCache & $74$ & $\mathbf{0.74}$ & $0.99$ & $0.88$ & $0.59$ & $0.47$ & $0.62$ & $\mathbf{82.8}$ \\
\rowcolor{gray!15}
\mosaik{}~(ours) + TaylorSeer & $74$ & $\mathbf{0.74}$ & $0.99$ & $0.87$ & $0.58$ & $0.47$ & $0.63$ & $\mathbf{82.8}$ \\
\bottomrule
\end{tabular}
\caption{\textbf{\mosaik{} vs.\ Uniform Patchification.} TR\,(\%) is the token reduction relative to $p16$ ($4096$ tokens). All rows use the same Stage~2 checkpoint. Bold marks the best GenEval and DPG score within each compute tier.}
\label{tab:mosaik_vs_uniform}
\end{table*}

\paragraph{One Damage Score Suffices for Both Refinements.}
Across $11{,}600$ denoising states from $232$ trajectories covering $120$ prompts, we measure for every region the error reduction from refining $p64$ to $p32$ and $p32$ to $p16$, and compare these gains with the shared damage score. The shared score closely matches both transition-specific rankings, with mean within-state Spearman correlations of $0.928$ and $0.869$ and $95\%$ prompt-cluster bootstrap confidence intervals of $[0.922,0.933]$ and $[0.856,0.881]$. Ranking all eligible refinements by gain per added token, the shared and transition-specific orderings remain strongly correlated at $\rho=0.856$ with a $95\%$ confidence interval of $[0.834,0.876]$. Replacing the shared score with the measured transition-specific gains reduces the remaining regional error relative to the $p16$ teacher by only $3.2\%$ to $4.7\%$ across token budgets of $15\%$ to $55\%$.

\section{Further Analysis}

\subsection{From FLOPs to Wall-Clock}
\label{sec:flop_accounting}
\label{sec:app_latency}

\paragraph{Whole-Denoiser Accounting.} To prevent overstated efficiency gains, we report FLOPs over the entire denoiser rather than over the dynamic components alone, as in Table~1 of the main paper. Table~\ref{tab:flops_anatomy} separates the two. The four image-side DiT modules scale with patch size and account for the whole reduction, whereas the text stream, the token-embedding, output-head and adaLN blocks, and the PiT pixel decoder do not: the decoder emits the full $1024\times1024$ image irrespective of the token layout. This fixed block, $21\%$ of a $p16$ step, caps any purely spatial method at a $75\%$ reduction. Our accounting is therefore conservative: a mixed layout averaging ${\sim}1024$ tokens is a $60\%$ whole-denoiser reduction, but would read as $79\%$ against the image stream alone. That accounting also charges \mosaik{} for the conditioning upsampler that lifts coarse tokens back onto the $p16$ grid the decoder requires, $19$ GFLOPs at $p32$ and $p64$ and none at $p16$.

\paragraph{The FLOPs Savings Hold at Runtime.}
We measure end-to-end latency on a single consumer-level GPU at $1024$px ($50$ steps, batch one), with the adapter merged into the base weights so \mosaik{} runs as one fused forward pass. Fusing the adapter does not inflate the $p16$ baseline ($21.78$ against $21.75$\,s). The overheads of varying patch size (token packing, positional remapping, layout construction) do not consume the savings: at equal token count an irregular layout is as cheap as a uniform one ($114.0$ against $113.2$\,ms). The damage-guided predictor and allocator add $2.0$\,ms per step, and packing stay at or below $1.0$\,s in every configuration (Fig.~\ref{fig:app_latency}). \mosaik{} is faster than uniform $p16$ at every budget, reaching $2.59\times$ at the maximal one: $8.4$\,s against $21.8$\,s, while peak memory stays flat at $9.7$\,GB. \mosaik{} is therefore a latency and throughput optimization rather than a memory one.

\paragraph{Where the Residual Gap Goes.} At the maximal budget the DiT blocks accelerate $5.40\times$ while the fixed block of Table~\ref{tab:flops_anatomy} becomes a hard wall-clock floor (Fig.~\ref{fig:app_latency}): of the $8.4$\,s, $3.1$\,s is DiT blocks, $4.4$\,s is the decoder, and $0.94$\,s is packing, remapping, and scheduling overhead. On consumer hardware the decoder, rather than memory bandwidth, is the binding constraint.

\subsection{Quality Under Compression}
\label{sec:app_quality}
\label{sec:app_ladders}

\paragraph{Limits of Uniform Patchification.}
Uniform patch scaling offers only three operating points, and the step between them is large. All uniform baselines in Table~\ref{tab:mosaik_vs_uniform} run the same Stage~2 checkpoint as \mosaik{} and differ from it only in that the layout is fixed instead of heterogeneous, so any difference between those rows is attributable to the layout alone. At the aggressive tier, uniform $p32$ is a strong configuration. It is also an isolated one: transitioning to the next discrete scale $p64$, where GenEval falls to $0.65$ and DPG to $80.2$, against $0.73$ and $83.2$ for uniform $p32$. The failure is compositional rather than per-object. Single-object accuracy is still $0.97$ at $p64$, while counting and attribute binding fall, because a $64$-pixel patch merges the spatial detail that separates one entity from another. The drop also shows that both benchmarks still resolve layout quality in the compression range these methods operate in, and are not saturated. \mosaik{} holds $0.74$ GenEval at both tiers by spending its fine patches where predicted damage is highest and its coarse budget elsewhere (Section~\ref{sec:app_alloc}).

\paragraph{Fidelity of the Generated Image.}
Prompt-alignment benchmarks score whether the requested objects, counts and attributes are present; they do not resolve texture. We therefore score every configuration with DISTS and LPIPS, two reference-based perceptual distances, against the image the full-compute PixelDiT produces from the same initial noise. The comparison is per image rather than distributional: it measures how far a configuration moves the specific image the uncompressed model would have produced. Table~\ref{tab:app_perceptual} reports the result. \mosaik{} at a $68\%$ reduction sits at $0.194$ DISTS and $0.392$ LPIPS, closer to the full-compute target than DDiT at a $14\%$ reduction ($0.217$ and $0.446$), and every \mosaik{} row is below every DDiT row on both metrics. The zero-coarsening row bounds what the coarsening itself costs. Stage~2 (Uniform $p16$) performs no coarsening and is already at $0.146$ DISTS, because the adaptation training has moved the weights away from the original model; \mosaik{} at a $52\%$ reduction is at $0.163$. At a moderate budget, most of the measured distance is therefore the price of the adaptation training rather than of the heterogeneous layout, and the distance grows gradually from there as the budget rises. Composing a step cache extends the budget without moving the distance: \mosaik{} with TeaCache and with TaylorSeer, both at $80\%$, sit at $0.198/0.391$ and $0.194/0.392$, level with \mosaik{} alone at $68\%$. What a heterogeneous layout buys is the interval of budgets, not the single point inside it that uniform coarsening happens to reach.

\begin{table}[t]
\centering
\footnotesize
\setlength{\tabcolsep}{3pt}
\renewcommand{\arraystretch}{1.15}
\begin{tabular}{@{}lccc@{}}
\toprule
\textbf{Configuration} & \textbf{FLOPs Red. (\%)} & \textbf{DISTS\,$\downarrow$} & \textbf{LPIPS\,$\downarrow$} \\
\midrule
\multicolumn{4}{@{}l}{\textit{Homogeneous Patchification}} \\
Stage~2 (Uniform $p16$) & $0$ & $0.146$ & $0.349$ \\
Stage~2 (Uniform $p32$) & $63$ & $0.173$ & $0.371$ \\
Stage~2 (Uniform $p64$) & $75$ & $0.249$ & $0.446$ \\
\midrule
\multicolumn{4}{@{}l}{\textit{Temporal Patchification}} \\
DDiT & $14$ & $0.217$ & $0.446$ \\
DDiT & $21$ & $0.227$ & $0.453$ \\
DDiT & $30$ & $0.237$ & $0.460$ \\
DDiT & $58$ & $0.258$ & $0.485$ \\
\midrule
\multicolumn{4}{@{}l}{\textit{Heterogeneous Patchification}} \\
\rowcolor{gray!15}
\mosaik{} & $52$ & $0.163$ & $0.374$ \\
\rowcolor{gray!15}
\mosaik{} & $62$ & $0.178$ & $0.381$ \\
\rowcolor{gray!15}
\mosaik{} & $68$ & $0.194$ & $0.392$ \\
\rowcolor{gray!15}
\mosaik{} + TeaCache & $80$ & $0.198$ & $0.391$ \\
\rowcolor{gray!15}
\mosaik{} + TaylorSeer & $80$ & $0.194$ & $0.392$ \\
\bottomrule
\end{tabular}
\caption{\textbf{Perceptual Distance to Full-Compute Baseline.} DISTS and LPIPS are measured against the original PixelDiT image generated from the same initial noise. FLOPs Red.\ is the whole-denoiser reduction.}
\label{tab:app_perceptual}
\end{table}

\subsection{Spatial Headroom in PDM}
\label{sec:app_why_pdm}
\paragraph{Matched Coarsening Protocol.}
We compare the perceptual cost of coarsening a VAE latent against the cost of coarsening native pixels on $n=500$ real images from DIV2K and FFHQ. This is a data-side measurement: no generator, checkpoint or training recipe enters it. Coarsening is $4\times$ area-pooling followed by bilinear upsampling, applied to each representation so that both sides are matched to the same effective pixel footprint per pooled unit and to the same post-coarsening token count. The denominator is the LPIPS damage from native pixel coarsening, measured against the original un-coarsened image. The numerator is the LPIPS damage from latent coarsening, measured against the VAE's own un-coarsened reconstruction rather than against the original, so that the VAE's inherent reconstruction error is excluded and only the coarsening is charged. That choice favours the latent side, since a latent diffusion model pays the reconstruction error in any case.

\begin{figure}[!t]
\centering
\includegraphics[width=\linewidth]{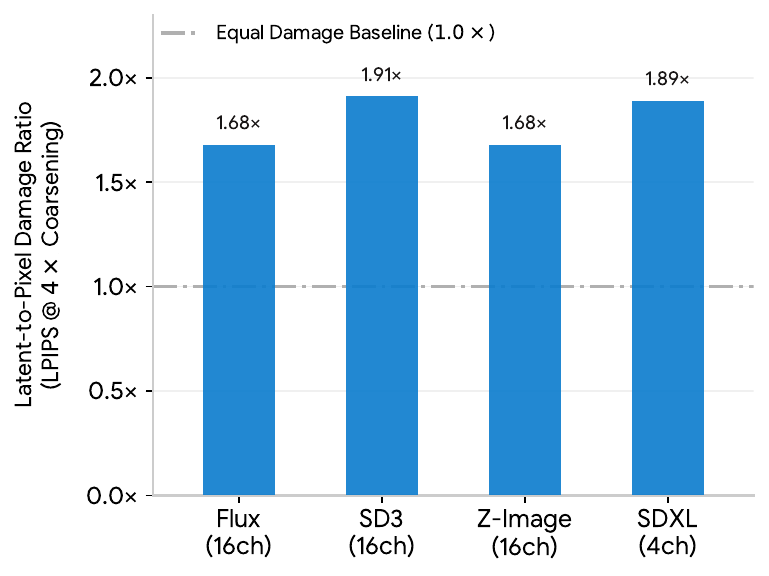}
\caption{\textbf{Latent-to-Pixel Coarsening Damage Ratio.} Each bar is the LPIPS damage from coarsening a VAE latent divided by the damage from coarsening native pixels at a matched factor. The dashed line marks the equal-damage reference at 1.0.}
\label{fig:vae_headroom}
\end{figure}

\paragraph{Preservation of Spatial Headroom.}
Coarsening the latent costs between $1.68\times$ and $1.91\times$ more perceptual damage than coarsening native pixels at a matched factor, for every autoencoder tested (Fig.~\ref{fig:vae_headroom}). The gap holds for a 4-channel autoencoder (SDXL) as well as three 16-channel ones (Flux, SD3, Z-Image), so it is not specific to one latent channel width or one vendor's autoencoder recipe. The reason is that the encoder has already spent the image's spatial redundancy once, on a fixed grid and identically everywhere, before generation begins: by the time a latent exists, little slack remains for a second pooling step to absorb, whereas the same operation applied to raw pixels still finds redundancy to consume. Adaptive coarsening can only pay off where redundancy is still available to spend, which is why \mosaik{} operates in pixel space.

\begin{table*}[t]
\centering
\footnotesize
\setlength{\tabcolsep}{6pt}
\renewcommand{\arraystretch}{1.15}
\begin{tabular}{@{}lcccc@{}}
\toprule
& \multicolumn{2}{c}{\textbf{Aggressive}} & \multicolumn{2}{c}{\textbf{Maximal}} \\
\cmidrule(lr){2-3} \cmidrule(l){4-5}
\textbf{Allocator Score} & \textbf{GenEval} & \textbf{95\% CI ($\Delta$)} & \textbf{GenEval} & \textbf{95\% CI ($\Delta$)} \\
\midrule
\multicolumn{5}{@{}l}{\textit{Compute Ceiling (Costly Oracle)}} \\
Measured Oracle & $0.7430$ & $[-0.61, +2.05]$ & $0.7345$ & $[-1.40, +1.09]$ \\
\midrule
\multicolumn{5}{@{}l}{\textit{Control Allocators}} \\
Content-blind (Random) & $0.7118$ & $[-3.97, -0.89]$ & $0.7026$ & $[-4.90, -1.84]$ \\
Inverted Damage & $0.6586$ & $[-9.66, -5.86]$ & $0.6628$ & $[-9.14, -5.53]$ \\
\midrule
\rowcolor{gray!15} \textsc{MOSAIK} (ours) & $\mathbf{0.7360}$ & & $\mathbf{0.7361}$ & \\
\bottomrule
\end{tabular}
\caption{\textbf{Ablation of Allocator Mechanisms.} Rows differ only in the score used to rank regions for refinement. Bracketed values are $95\%$ confidence intervals on the GenEval difference ($\Delta$) from \mosaik{}, in points.}
\label{tab:mosaik_ablation}
\end{table*}

\subsection{Ablation of the Damage-Guided Layout Predictor}
\label{sec:app_predictor}
Every row of Table~\ref{tab:mosaik_ablation} shares the Stage~2 multi-patch generator, the damage-guided layout predictor that refines regions in decreasing order of score per added token until the budget is exhausted.

The ceiling in the table is the measured oracle: the same allocator driven by the exact damage instead of the prediction, which requires a uniform $p16$ and a $p64$ forward pass at every denoising step to measure the deviation before a layout is committed. It is a measurement device rather than a deployable allocator, since by construction it costs more per step than running the uncoarsened model, whereas the predictor and its allocator together add $2.0$\,ms per step (Section~\ref{sec:flop_accounting}). At neither budget does the oracle's advantage over the predictor differ significantly from zero: the confidence interval on the gap is consistent with no true advantage, and bounds it at no more than $+2.05$ points at the aggressive budget and $+1.09$ points at the maximal one.

The two control allocators together show that the predicted damage ranking carries genuine information. The first, content-blind allocation, ignores the ranking entirely and serves as the random-baseline reference line; \mosaik{} outperforms it by a statistically clear margin at both budgets. The second control inverts the ranking, sending coarse patches to the regions of highest predicted damage rather than the lowest, and this inverted allocator scores several points below random. If the ranking carried no information, inverting it would still perform like random in expectation, since inverting noise is still noise. Scoring reliably below random therefore shows that the predicted damage ranks regions by their true sensitivity to coarsening, and in the correct direction. \mosaik{} is also the only method in the table whose score stays unchanged across both budgets.

\section{Detailed Allocation Behavior and Methodology}
\label{sec:app_alloc}

\subsection{Token Allocation Metric and Mask Generation}
\label{sec:app_alloc_metric}
Every method in Fig.~5 is scored against region masks taken from one shared reference image: the uniform $p16$ image generated for that prompt from the same initial noise. A segmentation model~(Mask2Former) is applied to that reference image alone, and the resulting masks are reused unchanged for \mosaik{} and DDiT so the area shares are identical across methods. The alternative, segmenting each method on its own image, would reward failure: a policy that coarsened the subject away would shrink its own subject mask and raise its own score.

Relative allocation is the ratio of a region's token share to its area share, so it equals exactly $1.0\times$ whenever a region receives tokens in proportion to its area, which is guaranteed whenever the patch size is spatially uniform within a step, regardless of how it varies across steps. DDiT is exactly this case: it changes patch size only over time, applying one uniform patch size to the entire frame at each step, so every region receives tokens in exact proportion to its area at every step. DDiT's relative allocation of $1.0\times$ in every region is therefore not an empirical result but a direct mathematical consequence of being spatially uniform. The same identity applies to a uniform $p32$ layout, which is competitive on quality at moderate budgets (Section~\ref{sec:app_quality}) but is likewise pinned to $1.0\times$ everywhere, since it too never varies patch size across space and so cannot express region-dependent allocation. The three regions are read from the same $16\times16$ grid of cells on which the layout itself is decided, so relative-allocation values are directly comparable across methods.

\subsection{Spatial Allocation Dynamics Under Budget Constraints}
Fig.~5 covers FLOPs reductions of $20\%$, $50\%$, $60\%$, and $70\%$. Allocation is stratified inside the subject itself: at the $60\%$ reduction the subject boundary receives $2.17\times$ its area share and the subject interior $1.80\times$, while the background receives $0.8\times$. Predicted damage therefore varies across a single semantic object, and the layout is decided at a finer granularity than a subject-against-background split. Across the same sweep the fraction of the background assigned coarse patches rises from $22\%$ to $75\%$, which is where the tokens kept for the subject come from.

At the $70\%$ reduction the realized subject premium eases slightly, while the premium theoretically attainable from the ground-truth mask keeps rising across the sweep, from $1.33\times$ to $5.31\times$. The easing is therefore not the budget forbidding further concentration. Subject share is also not the quantity the allocator maximizes: it applies the refinement with the highest predicted damage per added token until the budget is spent. The easing does not appear as a loss in prompt adherence, since \mosaik{} holds $0.74$ on GenEval at that tier while uniform $p64$ falls to $0.65$ (Table~\ref{tab:mosaik_vs_uniform}).

\section{DDiT Reimplementation Details}
\label{sec:app_ddit}
DDiT is evaluated on the Stage~1 checkpoint from which Stage~2 is initialized (Section~\ref{sec:app_stages_details}): it selects one patch size per denoising step, so it only executes uniform $p16$, $p32$, and $p64$, the family Stage~1 is trained on~(similar to DDiT training). Table~\ref{tab:mosaik_vs_uniform} instead runs those uniform layouts on \mosaik{}'s own Stage~2 weights. Everything else is matched: PixelDiT backbone at $1024$px, rank-$32$ LoRA, Gemma-2 text encoder, $50$-step Euler sampler, guidance value, and fixed seeds.

The schedule follows the published form: third-order finite difference of the sample trajectory, aggregated spatially at percentile $\rho = 0.4$, thresholded at sensitivity $\tau$. Because $\tau$ sets a sensitivity and not a budget, we swept it and report each run's realized reduction from its own per-step token counts. \mosaik{}'s lead is larger at every tighter tier.

\begin{table}[h]
\centering
\small
\renewcommand{\arraystretch}{1.1}
\begin{tabular}{llccc}
\toprule
\textbf{FLOPs Red.~(\%)} & $\tau$ & \textbf{DDiT} & \textbf{\mosaik{}} & \textbf{Lead} \\
\midrule
21.2 & $0.002634$ & 0.6936 & 0.7385 & +4.5 \\
49.3 & $0.003026$ & 0.6650 & 0.7408 & +7.6 \\
61.5 & $0.003449$ & 0.6500 & 0.7360 & +8.6 \\
70.0 & $0.007304$ & 0.6546 & 0.7361 & +8.2 \\
\bottomrule
\end{tabular}
\caption{\textbf{DDiT Reimplementation.} $\tau$ is the schedule's sensitivity threshold and FLOPs Red.\ the reduction each run realized. DDiT and \mosaik{} columns are overall GenEval; Lead is \mosaik{} minus DDiT, in points.}
\label{tab:ddit_canonical}
\end{table}

\begin{figure*}[t]
\centering
\includegraphics[width=0.8\linewidth]{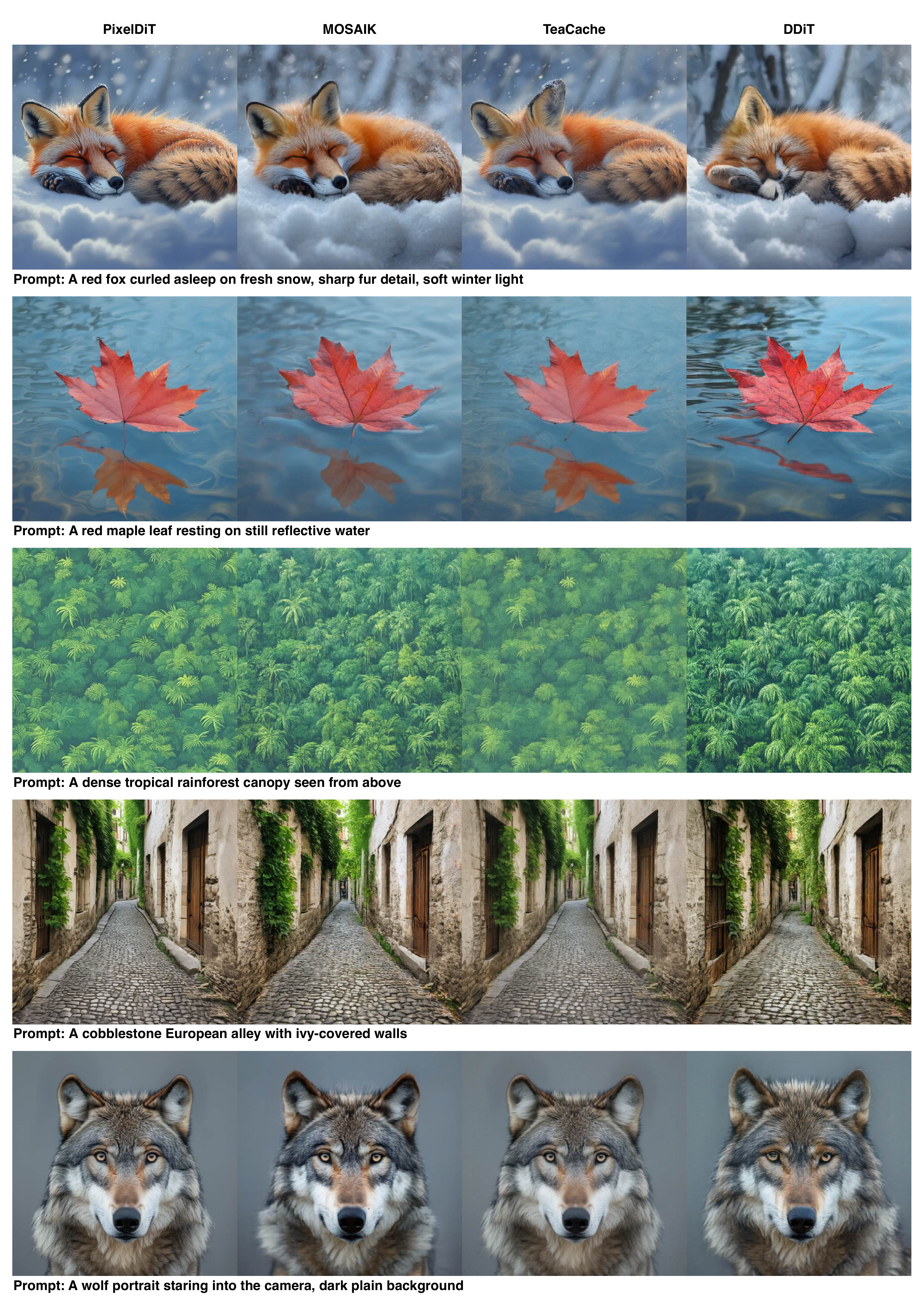}
\caption{\textbf{Additional Text-to-Image Generation Results.} The caption beneath each image shows the corresponding generation prompt. \mosaik{} against DDiT and TeaCache, all at a $70\%$ FLOPs reduction. The base model with the highest FLOPs is included as a reference.
}
\label{fig:vis-all-1}
\end{figure*}
\begin{figure*}[t]
\centering
\includegraphics[width=0.79\linewidth]{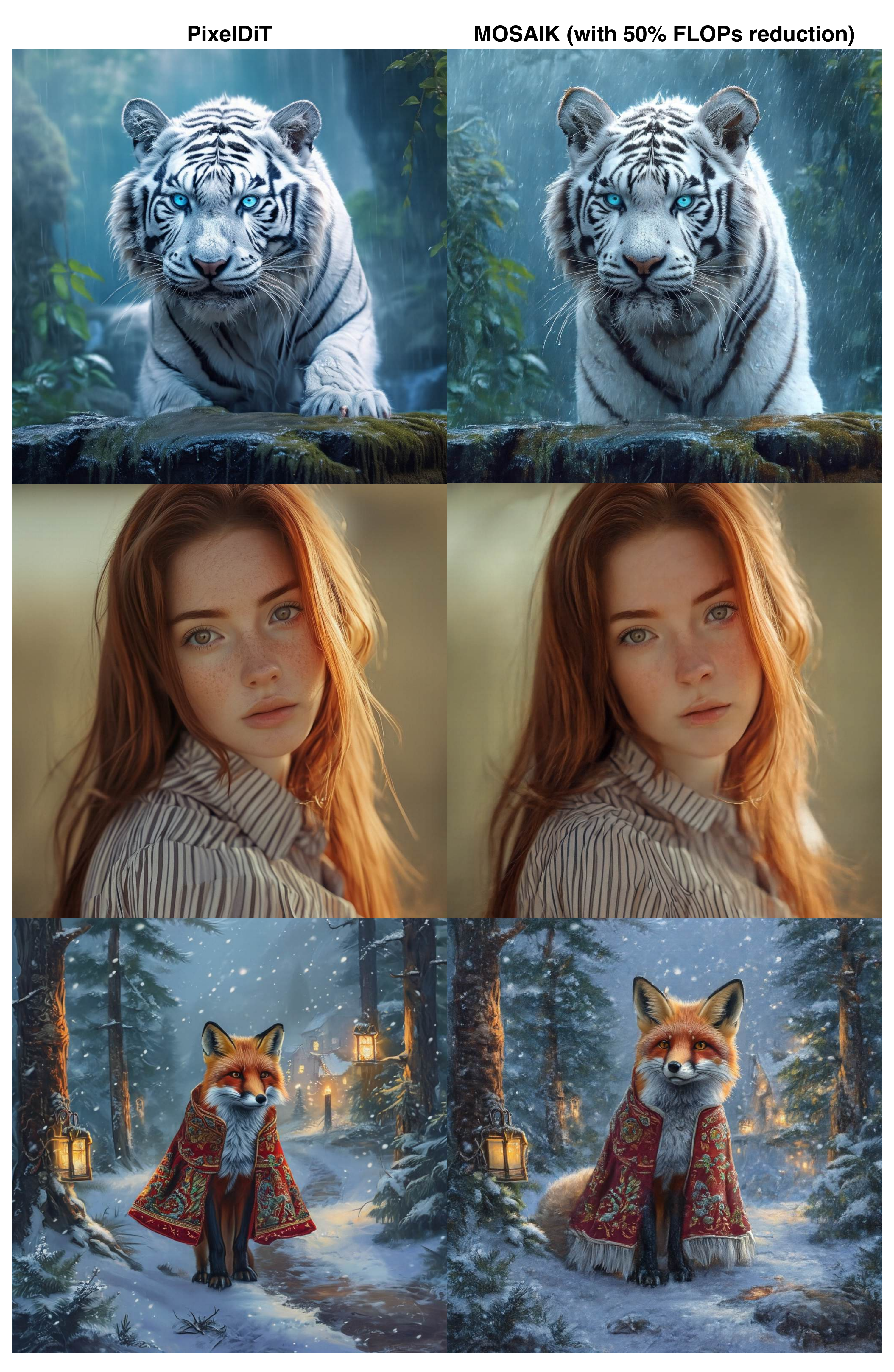}
\caption{\textbf{Additional Text-to-Image Generation Results.} The caption beneath each image shows the corresponding generation prompt. PixelDiT vs \mosaik{} at $50\%$ FLOPs reduction.
}
\label{fig:vis-all-2}
\end{figure*}

\end{document}